\documentclass[sigconf, nonacm]{acmart}

\usepackage{times}
\usepackage[utf8]{inputenc} 
\usepackage[T1]{fontenc}    
\usepackage{hyperref}       
\usepackage{url}            
\usepackage{booktabs}       
 \usepackage{amsfonts}        
\usepackage{nicefrac}       
\usepackage{microtype}      
\usepackage{xcolor}        
\usepackage{algorithm}
\usepackage{algorithmic}
\usepackage{times}
\usepackage{soul}
\usepackage[]{caption}
\usepackage{graphicx}
\usepackage{bbding}
\usepackage{amsmath}
\usepackage{amsthm}
\usepackage[switch]{lineno}
\usepackage{libertine}
\usepackage{pifont}
\usepackage{enumitem}
\usepackage{wrapfig}

\usepackage{multicol}
\usepackage{threeparttable}
\usepackage{indentfirst} 
\usepackage{cases}

\usepackage{graphicx}
\usepackage{epstopdf}
\usepackage{array}
\usepackage{pdflscape}
\usepackage{siunitx}
\usepackage{makecell}
\usepackage{multirow}
\usepackage{adjustbox}

\newtheorem{theorem}{Theorem}
\newtheorem{lemma}{Lemma}

\begin{document}

\title[The Puppeteer's Attack for Fine-Grained Control in Ranking-Based Federated Learning]{Beyond Denial-of-Service: The Puppeteer's Attack for Fine-Grained Control in Ranking-Based Federated Learning}

\author{Zhihao Chen}
\affiliation{%
  \institution{Fujian Normal University}
  \city{Fuzhou}
  \state{Fujian}
  \country{China}
}
\email{zhihao_edu@163.com}

\author{Zirui Gong}
\affiliation{%
  \institution{Griffith University}
  \city{Gold Coast}
  \state{Queensland}
  \country{Australia}
}
\email{zirui.gong@griffithuni.edu.au}

\author{Jianting Ning}
\affiliation{%
  \institution{Fujian Normal University}
  \city{Fuzhou}
  \state{Fujian}
  \country{China}
}
\authornote{Corresponding author.}
\email{jtning88@gmail.com}

\author{Yanjun Zhang}
\affiliation{%
  \institution{Griffith University}
  \city{Brisbane}
  \state{Queensland}
  \country{Australia}
}
\email{yanjun.zhang@griffith.edu.au}

\author{Leo Yu Zhang}
\affiliation{%
  \institution{Griffith University}
  \city{Gold Coast}
  \state{Queensland}
  \country{Australia}
}
\email{leo.zhang@griffith.edu.au}


\begin{abstract}
Federated Rank Learning (FRL) is a promising Federated Learning (FL) paradigm designed to be resilient against model poisoning attacks due to its discrete, ranking-based update mechanism. Unlike traditional FL methods that rely on model updates, FRL leverages discrete rankings as a communication parameter between clients and the server. This approach significantly reduces communication costs and limits an adversary's ability to scale or optimize malicious updates in the continuous space, thereby enhancing its robustness.
This makes FRL particularly appealing for applications where system security and data privacy are crucial, such as web-based auction and bidding platforms.
While FRL substantially reduces the attack surface, we demonstrate that it remains vulnerable to a new class of local model poisoning attack, i.e., fine-grained control attacks.

We introduce the \underline{E}dge \underline{C}ontrol \underline{A}ttack (ECA), the first fine-grained control attack tailored to ranking-based FL frameworks. Unlike conventional denial-of-service (DoS) attacks that cause conspicuous disruptions, ECA enables an adversary to precisely degrade a competitor’s accuracy to any target level while maintaining a normal-looking convergence trajectory, thereby avoiding detection.
ECA operates in two stages: (i) identifying and manipulating Ascending and Descending Edges to align the global model with the target model, and (ii) widening the selection boundary gap to stabilize the global model at the target accuracy.
Extensive experiments across seven benchmark datasets and nine Byzantine-robust aggregation rules (AGRs) show that ECA achieves fine-grained accuracy control with an average error of only 0.224\%, outperforming the baseline by up to 17×. Our findings highlight the need for stronger defenses against advanced poisoning attacks.
We release our code at \href{https://github.com/Chenzh0205/ECA}{https://github.com/Chenzh0205/ECA} to foster further research in this area.
\end{abstract}








\maketitle

\section{Introduction}
\label{introduction}
\begin{figure}[t!]
    \centering
    \includegraphics[width=\linewidth]{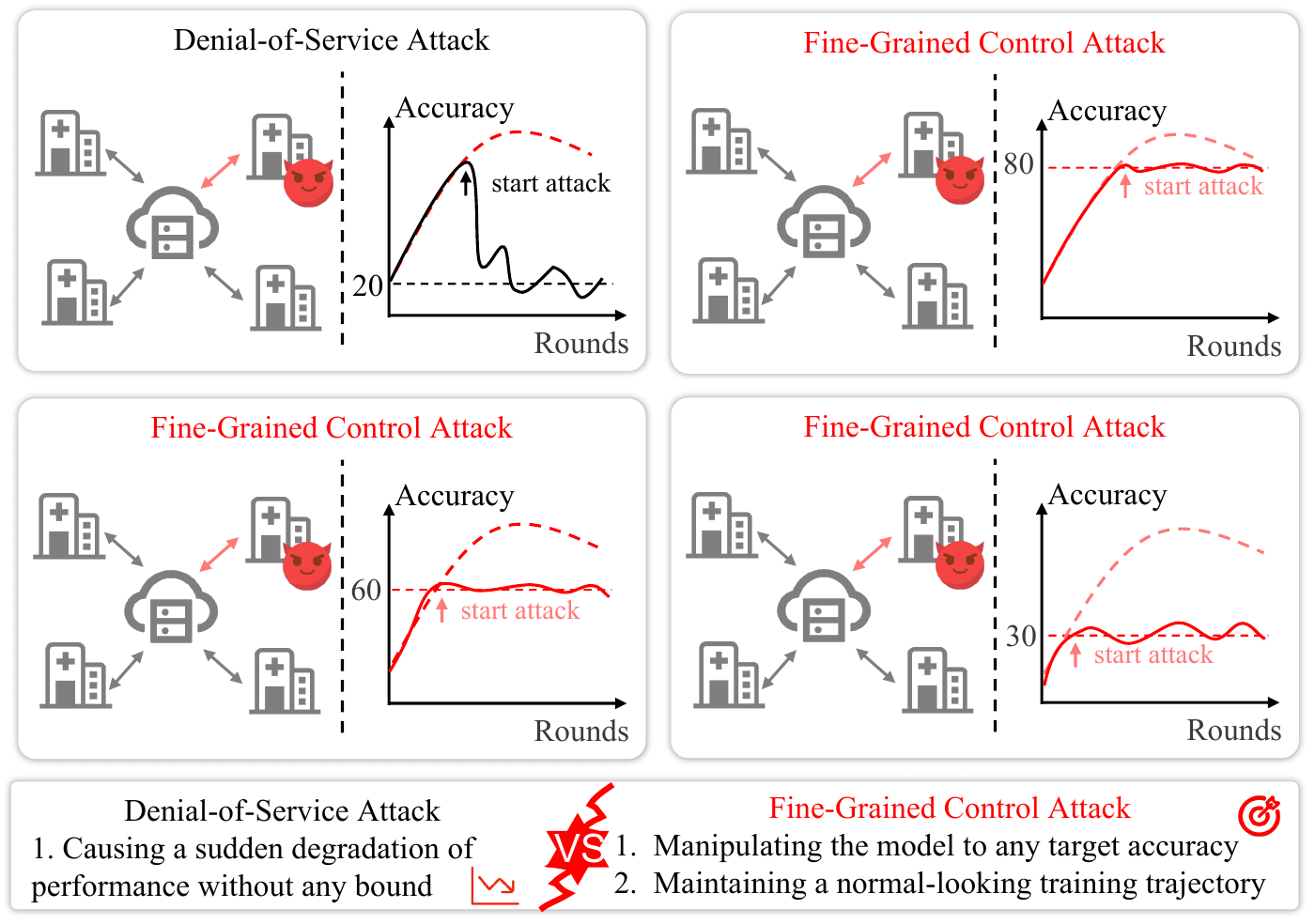}
    \caption{Illustration of the strategic advantages of a fine-grained control attack over a Denial-of-Service (DoS) attack.}
    \Description{Plot.}
    \label{fig:intro_overall}
    \vspace{-0.2cm}  
\end{figure}
Web-based auction and bidding platforms \citep{yang2023federated,dutting2021optimal,cui2024auction} are systems where multiple advertisers compete in real time to display ads to users, with the highest bidders winning ad placement opportunities. These platforms rely heavily on machine learning models to predict click-through rates, optimize bid strategies, and detect fraudulent behavior. The models are trained on vast amounts of user interaction data, such as browsing history, past ad interactions, and contextual signals, to maximize platform revenue and user engagement. However, centralizing such sensitive data raises  privacy and security concerns, as it could expose personal user information or proprietary bidding strategies.
Federated learning (FL) \citep{mcmahan2017communication} offers a promising solution to this challenge.
FL is a distributed machine learning paradigm where multiple clients collaborate to train a shared model without exposing their local data. Each client trains a local model on its dataset and sends updates to a central server, which aggregates them to form a global model, then redistributes it for further training.

However, the distributed nature of FL makes it particularly vulnerable to local model poisoning attacks \cite{zhang2023denial,baruch2019little,blanchard2017machine,shejwalkar2021manipulating,fang2020local,jagielski2018manipulating,bagdasaryan2020backdoor,bhagoji2019analyzing,SPzirui,DLiWNHXZW24}. One type is the Denial-of-Service (DoS) attack, where the goal is to maximally degrade the global model’s accuracy. Although effective, DoS attacks have two major limitations. First, they only aim to maximize the global model's accuracy drop, so the eventual level of degradation is unpredictable. Second, they typically produce a sudden and erratic accuracy drop, making the attack conspicuous (see Fig. \ref{fig:intro_overall}). Recently, a new attack paradigm, i.e., fine-grained control attack has been proposed~\cite{zhang2023denial,pan2025}. It aims to (i) manipulate the global model’s accuracy to \textbf{any} target value and (ii) maintain a plausible training trajectory that mimics normal convergence. As illustrated in Fig. \ref{fig:intro_overall}, this allows the attacker to steer the model to various target accuracies (e.g., 80\%, 60\%, or 30\%) while preserving a natural-looking training curve. This affords the attacker more precise, covert, and practical control over global model performance.
 
To defend against poisoning attacks in FL, various defense methods have been proposed. One significant category is Byzantine-robust aggregation rules (AGRs) \cite{cao2020fltrust,fang2022aflguard,mhamdi2018hidden,nguyen2022flame,sun2019can,xu2024robust,yin2018byzantine,zhang2022fldetector,blanchard2017machine,TIFSzirui}, which aim to detect and exclude malicious client updates. However, these AGRs can be bypassed by adaptive attacks, in which adversaries carefully craft their updates to look similar to normal ones while still causing significant damage.
Recent advancements suggest redesigning the FL framework for inherent robustness, with ranking-based FL \citep{mozaffari2023every,seoprism,zhao2024fefl,guo2025exploring,guastellasparsyfed} being a state-of-the-art approach. This method replaces gradient exchange with discrete edge rankings and differs from standard FL in two key ways. First, it uses discrete rankings as the communication parameters between clients and the server, where each ranking indicates the relative importance of edges in the network. Second, on the server side, it employs majority voting to determine the global ranking and then applies the Lottery Ticket Hypothesis (LTH) \citep{frankle2019lottery} to select only the top k\% most important edges to form a subnetwork, which can achieve the same performance as the full supernetwork. By avoiding gradient exchange and reducing the attack surface, ranking-based FL significantly improves robustness and makes existing gradient-based attacks ineffective. 

\noindent\textbf{Our work:} 
In this work, we demonstrate that even when the victim system employs the ranking-based FL, i.e., FRL, to train its models, it is still feasible to launch effective fine-grained control attacks by introducing a small amount of malicious clients into the system. 
Specifically, the attacker aims to (i) manipulate the model's accuracy to any desired target value, and (ii) ensure the global model maintains a normal convergence trend throughout the training process. This dual capability of achieving precise control while maintaining stealth makes the attack highly practical in real-world scenarios.  Unlike overt DoS attacks that trigger obvious alarms, our fine-grained approach allows the adversary to subtly dictate the model's performance, making it a more covert and realistic threat.

To achieve the goal, we introduce the Edge Control Attack (ECA), the first fine-grained control attack that targets the state-of-the-art (SOTA) robust FL framework, i.e., FRL.
To execute the attack, we propose a two-stage algorithm. In the first stage, we identify the set of Ascending and Descending Edges that represent the edge difference between the current global model and the target model. These edges are then manipulated to positions that maximize the attack’s impact.
In the second stage, we aim to fix the global model’s accuracy at the target value while simulating a normal convergence trend. To achieve this, we deliberately increase the gap required for the global model to change, preventing benign updates from reversing our manipulation. This stabilizes the model at the target state, ensuring that its accuracy cannot increase or decrease further.

We evaluate our attack on seven benchmark datasets under nine robust aggregation rules (AGRs), demonstrating our attack can achieve a high level of control across all combinations of datasets, target accuracy, and defenses, with an overall 0.224\% control error, and outperforms another attack by 17$\times$.
Our key contributions are highlighted as follows: 
\begin{itemize}[leftmargin=*]
    \item We identify a new strategic vulnerability in ranking-based FL paradigms that extends beyond simple disruption. We demonstrate that an adversary can exploit the discrete update mechanism not just to damage a model, but to precisely steer its accuracy to any specific, attacker-chosen level, creating a potent threat for competitive scenarios like commercial bidding.
    \item We design the Edge Control Attack (ECA), the first attack method specifically tailored to achieve fine-grained control in FRL.  We propose a novel algorithm that is more practical in real-world scenarios, enabling attackers to achieve dual objectives: (i) manipulating the model to any desired accuracy, and (ii) maintaining a normal-looking training trajectory to evade detection. We provide a theoretical analysis that formally establishes the conditions under which this precise control is feasible and highly probable.
    \item We conduct extensive experiments demonstrating that ECA poses a significant and practical threat.  Our results show that ECA can maintain control with an exceptionally low error (0.224\%) across diverse settings, significantly outperforming relevant baselines. 
\end{itemize}
 
\section{Background and Related Work}
\label{background}
\subsection{Federated Rank Learning~(FRL) and Poisoning Attack in FRL}
FRL~\citep{mozaffari2023every} is a SOTA FL paradigm that operates in a discrete space by leveraging the Lottery Ticket Hypothesis \citep{frankle2019lottery} and utilizing the edge-popup (EP) algorithm~\citep{ramanujan2020s} to identify sparse subnetworks capable of achieving comparable performance to dense models. Specifically, the FRL has the following processes.

\noindent\textbf{Server Initialization:} The server uses a random seed to initialize the weights $\theta_w$ and scores $\theta_s$ for each edge within the \textit{global supernetwork}. During training, only the scores \(\theta_s\) are updated. The server then sorts $ \ theta_s$ in ascending order to create the global ranking $R_g$, where each value represents the ID of the edge and a higher rank indicates greater importance. Finally, the server sends the global ranking $R_g$ to the randomly selected $U$ clients.

\noindent\textbf{Clients Training:} Each client reconstructs the score $\theta_s$ using the received global ranking $R_g$ and the local data $D_u$. The scores $\theta_s$ for each edge are then updated and ranked using the EP algorithm and local data to obtain the updated ranking. If an edge $e$ in the \textit{subnetwork} reduces the training loss (e.g., cross-entropy), then the score of this edge will increase. Finally, clients send the updated rankings to the server for aggregation.

\noindent\textbf{Server Aggregation:} Upon receiving the rankings $R_u$ from $U$ clients, the server performs majority voting (MV) to get the aggregated global ranking. Specifically, it has the following steps: i) the server first aligns the \textit{importance scores} $I_u$ of all edges across clients based on their edge IDs; ii) then it performs a dimension-wise summation across clients, and calculates the \textit{aggregated importance scores} for each edge, i.e., $S = \sum_{u=1}^{U} I_u$; iii) finally, the edges are rearranged based on the \textit{aggregated importance scores} to produce the new global ranking $R_g^{t+1}$. For example, suppose two clients upload rankings $R_1 = [0,4,2,3,5,1],~R_2 = [0,1,2,5,4,3]$, where $e_0$ is the least important edge (at the first index). Then the server computes the \textit{importance score} vector as $I_1 = [0,5,2,3,1,4], ~I_2 = [0,1,2,5,4,3]$. After aggregating scores, the server calculates the \textit{aggregated importance scores}, i.e., $ S =[0,6,4,8,5,7].$ Based on the \textit{aggregated importance scores} of each edge, the server gets the aggregated global ranking $R_g^{t+1} = [0,2,4,1,5,3].$

Finally, the global ranking $R_g^{t+1}$ is transformed into a binary supermask $\textbf{M}$, where the top $k\%$ edges are set to 1, and the remaining edges are set to 0, then reordered to align with the original edge IDs. For example, if the global ranking is $R_g^{t+1} = [0,2,4,1,5,3]$ and k=50\%, edges $e_1$, $e_5$, $e_3$ will be selected for \textit{subnetwork}, therefore, the corresponding mask is $\textbf{M}[R_g^{t+1}] = [0,1,0,1,0,1]$. In summary, the goal of FRL is to find a global ranking $R_g$, such that the resulting \textit{subnetwork} $\theta_w \odot \textbf{M}$ minimizes the average loss of all clients. Formally, we have
\begin{align}
&\min_{R_{g}} F(\theta^{w},R_{g}) = \min_{R_{g}} \sum_{u = 1}^{U} \lambda_{u} L_{u}(\theta^{w} \odot \textbf{M}),\\ \nonumber
&\text{s.t.}~ \textbf{M}[R_{g}[e]] = \begin{cases} 0 & \text{if~} e < t, \\ 1 & \text{if~} e \geq t, \end{cases}
\end{align}
where $U$ denotes the number of clients selected in training, $\theta^{w}$ denotes the global model weights. $L_u$ and $\lambda_{u}$ denote the loss function and weight at aggregation for the $u^{th}$ client, respectively, with $\lambda_{u}=1/U$. The index of the selection boundary for the \textit{subnetwork} is $t = \lfloor \left(1-k\right)n\rfloor $, where n is the number of edges in each layer and typically we use the \textit{subnetwork} size $k=50\%$. 

While FRL's discrete nature and majority voting mechanism were designed to enhance robustness, recent work has demonstrated that it is not immune to poisoning. Specifically, the Vulnerable Edge Manipulation (VEM) attack~\citep{SPzirui} revealed that FRL is susceptible to DoS attacks. The core finding of VEM is that not all edges are equally robust; those near the subnetwork selection boundary are particularly ``vulnerable" to manipulation. By identifying this set of vulnerable edges, VEM employs an optimization-based method (Gumbel-Sinkhorn) to craft malicious rankings that maximize performance degradation, successfully undermining the global model. 

However, the threat model addressed by VEM is confined to a DoS objective, which reveals several critical limitations.  First, its strategy of maximizing disruption is fundamentally incompatible with the goal of precise control.  VEM's optimization-based approach is designed for a one-way, unbounded degradation, making it incapable of reliably steering a model to a specific, predetermined accuracy level.  It lacks the mechanisms to prevent the model's performance from collapsing far below a target, a crucial requirement for a control task.  Second, this pursuit of maximum damage inherently sacrifices stealth. A significant drop in model accuracy is a telltale sign of attack activity, making the compromised system highly detectable. For attackers seeking to maintain a low profile, such attacks are both difficult to execute and impractical.

\subsection{Fine-Grained Control Attacks in FL}
Beyond simple DoS, a more advanced threat is the fine-grained control attack, where an adversary aims to manipulate a model's performance to a specific level of accuracy. This type of attack is stealthier and can be more damaging in competitive commercial scenarios than a DoS attack. 
However, existing fine-grained control attacks are designed for traditional FL and do not apply to FRL.

For instance, two prominent methods, the Flexible Model Poisoning Attack (FMPA) \citep{zhang2023denial} and the Federated Learning Sliding Attack (FedSA) \citep{pan2025}, both operate in a continuous parameter space. FMPA's core strategy is model replacement: it first predicts the likely state of the benign global model for the next round, and then crafts a malicious update intended to overwrite it with a poisoned version that exhibits a specific low accuracy. FedSA, on the other hand, approaches the problem from a control theory perspective. It models the entire FL training process as a dynamic system and defines an ``error" as the difference between the current model's accuracy and the desired target accuracy. The attacker then calculates a malicious update using a specifically designed ``control law". This isn't a fixed poisoned model, but rather a dynamic adjustment calculated each round, acting to continuously correct the global model's trajectory, actively steering it towards the compromised state and theoretically guaranteeing its convergence to the attacker's precise target.

Crucially, since both methods achieve control by manipulating continuous weights or gradients, they are fundamentally incompatible with FRL, which functions within a discrete ranking space where such perturbations are meaningless. This reveals a critical research gap: no existing methods are capable of performing fine-grained accuracy control on ranking-based FL systems.

\section{Edge Control Attack (ECA)}
\label{ECA}

\subsection{Threat Model}
\noindent\textbf{Attacker’s objective.} 
In this work, we move beyond the traditional DoS threat model \citep{baruch2019little,fang2020local,shejwalkar2021manipulating}. Instead of aiming for maximum disruption, the adversary pursues a more sophisticated, goal-oriented objective inspired by competitive scenarios like commercial bidding. The attacker's primary goal is to control the competitor's global model precisely, steering its final accuracy to a specific, \textbf{any} desired value ($\tau$) that is strategically advantageous (e.g., marginally lower than their own model's accuracy). 
This primary goal is coupled with a critical secondary objective: stealth. The attack must be executed in a way that the model's training trajectory appears normal to an outside observer. This means avoiding the conspicuous, abrupt performance drops characteristic of DoS attacks. The attacker's dual objectives are therefore: (i) to ensure the global model converges to the precise target accuracy ($\tau$), and (ii) to maintain a convergence path that mimics a natural, albeit slower or less effective, training process to evade detection.

\noindent\textbf{Attacker’s capability.} 
We assume that the attacker can control $m$ malicious clients out of $U$ total clients per round. In the context of our threat model, these could be sleeper agents introduced into a competitor's team during an FL competition. Following the common assumption in previous studies \citep{fang2020local,shejwalkar2021manipulating}, we set the proportion of malicious clients to 20\% in our main experiments.

\noindent\textbf{Attacker’s knowledge.} We consider update-agnostic settings, where the attacker has no knowledge of benign client updates but uses the clean dataset of malicious clients and the information received from the global model to generate malicious updates.

\subsection{Motivation} 
In FRL, model accuracy is determined by the mask $\textbf{M}$, which controls the selection of edges that form the subnetwork. Therefore, to achieve the goal of steering the global model accuracy to any desired level while maintaining the normal training convergence trend, we focus on designing the malicious mask. 
Specifically, we aim to minimize the mask difference between the poisoned global ranking $\hat{R}_g$ and a target ranking $R_\tau$.
This leads to the following objective function:
\begin{align}
    \label{eq:formulation}
    &\min ~\left|\textbf{M}[R_\tau]-\textbf{M}[\hat{R}_g]\right|, \\ \nonumber
    &\text{s.t.}~\hat{R}_g = \text{MV}(\hat{R}_1, \ldots, \hat{R}_m, R_{m+1}, \ldots, R_{U}),
\end{align} 
where $R_\tau$ is the target global ranking with a test accuracy of $\tau$, $\hat{R}_g$ is the poisoned global ranking, $\textbf{M}[R_\tau]$ and $\textbf{M}[\hat{R}_g]$ are the masks of the two rankings, MV() is the majority voting process, $\hat{R}_u$ (for $u \in m$) are the malicious rankings, and $R_u$ (for $u \in [m+1,U]$) are the benign rankings.
However, achieving this objective is challenging and can not be solved by existing methods. Existing fine-grained control attack \cite{zhang2023denial} usually adds an optimized perturbation on the malicious gradient, aiming to cancel out further changes in the global gradient.
However, this mechanism relies on adding a small noise to gradient updates, which will introduce floating-point values into the ranking updates, violating the format constraint of FRL.
Alternatively, one might consider applying a similar optimized perturbation to the client-side scores $\theta_s$ to mimic gradient-level manipulation in traditional FL. However, in FRL, clients have no access to the global score, which is also a key factor contributing to the robustness of FRL. This inherent property makes such optimization infeasible.
Although VEM proposes to use the Gumbel-Sinkhorn method to solve optimization problems in discrete spaces \cite{SPzirui}, the stochastic and uncertain nature of permutation-based optimization prevents it from precisely steering the model toward any target accuracy.

\subsection{Overall of ECA}
\begin{figure}[t!]
    \centering
    \includegraphics[width=\columnwidth]{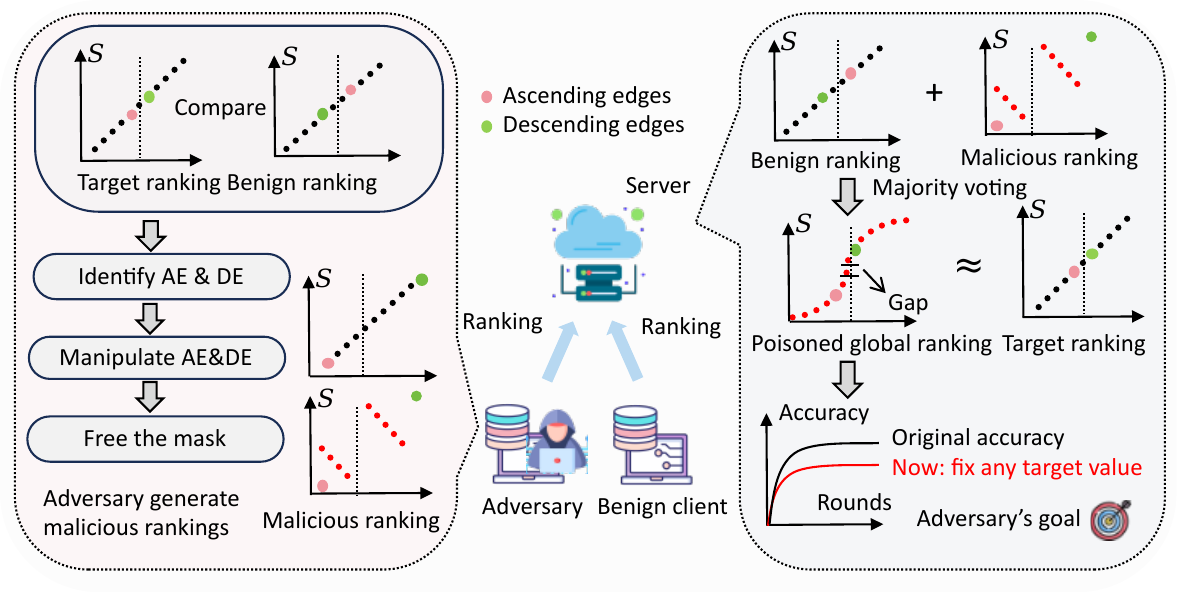}
    \caption{Overview of the ECA attack. The left shows the adversary generates malicious updates by aligning the global model mask with the target and freezing it via a widened boundary gap. The right shows the attack’s effect on the global model, enabling precise and stable control of its accuracy.}
    \Description{Plot.}
    \label{fig:overall}
    \vspace{-0.3cm}
\end{figure}
To solve the problem, we propose a deterministic two-stage algorithmic construction based on permutation-based methods \citep{reiter1965discrete,deshwal2022bayesian,zhang2019bayesian} to directly find a feasible malicious ranking $\hat{R}$ that achieves our goal.
Our method, as shown in Fig. \ref{fig:overall}, consists of two main stages: Manipulate Ascending \& Descending Edges and Freeze the Mask.

In the first stage, the adversary aims to align $\textbf{M}[\hat{R}_g] $ with $\textbf{M}[R_\tau]$ by manipulating specific edges. To achieve this, it first identifies the mask difference between the target ranking ($R_{\tau}$) and the aggregated ranking of other $U-m$ benign clients ($\overline{R}$). These differing edges are classified into ascending edges (AE) and descending edges (DE).
The adversary then manipulates AE \& DE to generate intermediate malicious rankings ($\hat{R}'$) that offset these discrepancies (details see Section \ref{step1}).
In the second stage, the adversary aims to freeze the mask of $\hat{R}_g$ to prevent further updates.
To achieve this, it reverses the ranking on each side of the subnetwork boundary based on $\hat{R}'$, we refer to it as internal reverse in the following section. This process will reduce the importance scores of edges on the proximal left side of the boundary (outside the subnetwork) while increasing the importance scores of edges on the proximal right side (inside the subnetwork). Then during the majority voting stage, the generated malicious ranking will widen the boundary gap required for a mask change in $\hat{R}_g$, thereby ensuring continued alignment with the mask of target ranking $\textbf{M}[R_{\tau}]$ (details see Section \ref{step2}).

\subsection{Manipulate Ascending $\&$ Descending Edges}
\label{step1}
Since $\hat{R}_g \approx \text{MV}(\hat{R}, \overline{R})$, where $\hat{R}$ are malicious rankings and $\overline{R}$ is aggregated ranking of other $U-m$ benign clients, 
to achieve Eq. \ref{eq:formulation}, we first need to identify the mask different between $R_{\tau}$ and $\overline{R}$, and then use malicious rankings to offset this difference.

Assuming we have the aggregated rankings of other $U-m$ benign clients ($\overline{R}$), to identify the mask difference between  $R_{\tau}$ and $\overline{R}$, we first need to initialize the target ranking $R_\tau$.
Specifically, we assign $R_\tau=R_g^{t_0}$ (where $t_0$ is the first round when $ACC(R_g^{t_0}) \geq \tau$). And we use $\varepsilon$ to denote the initialization error between $ACC(R_g^{t_0})$ and $\tau$, i.e., $\varepsilon=|ACC(R_g^{t_0})-\tau|$.
During subsequent rounds, 
$R_\tau$ and $\varepsilon$ are dynamically updated to minimize $\varepsilon$. 
Specifically, 
if $|ACC(R_g^t)-\tau|<\varepsilon$, we assign $R_g^t$ to $R_{\tau}$, and update $\varepsilon= |ACC(R_g^{t})-\tau|$.

After determining $R_{\tau}$, we start to identify different edges where $\textbf{M}[R_\tau[e]] \ne \textbf{M}[\overline{R}[e]]$. 
Upon observation, we find two types of edges: Ascending Edge (AE) and Descending Edge (DE).
AE are those edges that are not selected by $R_\tau$, but selected by $\overline{R}$, defined as $\mathrm{\mathbb{I}_{AE}}=\{e\mid\textbf{M}[R_\tau[e]]=0 \mathrm{~and~} \textbf{M}[\overline{R}[e]]=1\}$.
Conversely, DE are those edges that are selected by $R_\tau$, but not selected by $\overline{R}$, defined as $\mathrm{\mathbb{I}_{AE}}=\{e\mid\textbf{M}[R_\tau[e]]=1 \mathrm{~and~} \textbf{M}[\overline{R}[e]]=0\}$.
Then to achieve Eq. \ref{eq:formulation} and align  $\textbf{M}[\hat{R}_g]$ with $\textbf{M}[{R}_{\tau}]$, we reduce the importance of AE and increase the importance of DE in our malicious rankings.
As illustrated in Fig.~\ref{fig:overall}, we move AE to the beginning of the $\overline{R}$ and move DE to the end of $\overline{R}$ to get intermediate malicious ranking $\hat{R}'$.

Now we illustrate two methods to estimate the aggregated rankings of other $U-m$ benign clients ($\overline{R}$).
\textbf{Historical estimation}, which assumes that the global model changes slowly over training rounds, allowing the ranking from the previous round ($R_g^{t-1}$) to approximate the current $\overline{R}^t$,
i.e.,
$\overline{R}^t \approx R_g^{t-1}$.
\textbf{Alternative estimation} uses malicious clients' local rankings to approximate the $\overline{R}$ for the current round, i.e., $\overline{R} \approx \text{MV}(R_1, \ldots, R_m)$.

\subsection{Freeze the Mask}
\label{step2}

After aligning the mask of poisoned global ranking $\hat{R}_g$ with target raking $R_{\tau}$, we aim to freeze the mask of $\hat{R}_g$ to prevent further updates.
However, In FRL, the mask near the subnetwork boundary is particularly sensitive to changes because the aggregated importance scores $S$ of edges on either side of the boundary are shown to be close. 
This sensitivity is a natural feature of FRL, as it enables dynamic selection of subnetwork edges to optimize global model accuracy.

To overcome this challenge and freeze the mask, we maliciously amplify the cost (i.e., the boundary gap) of edges that need to cross the subnetwork boundary.
We reverse the rankings on both sides of the boundary based on $\hat{R}'$ (except AE \& DE) to obtain the final malicious ranking $\hat{R}$. This approach reduces the importance score of edges on the proximal left side of the boundary (outside the subnetwork) while increases those on the proximal right side of the boundary (inside the subnetwork) as much as possible while ensuring that the edge selection results remain unchanged. Then after server-side majority voting, the global ranking forms a sigmoid-like distribution, where the boundary gap increases. 

Algorithm.~\ref{alg:ECA} describes our ECA process. The attack is triggered once the global model accuracy $ACC(R_g^t)$ surpasses the target threshold $\tau$. The attacker captures the current global ranking as an initial target ranking $R_\tau$ and dynamically updates it in subsequent rounds to find the ranking that most closely matches the target accuracy $\tau$ (lines 4-9). The main attack then commences (lines 10-18). In each round, the attacker first estimates the aggregated ranking of benign clients, $\overline{R}^t$ (line 12). By comparing $\overline{R}^t$ with the target $R_\tau$, it identifies the precise sets of AE/DE (line 13). To align the model mask with the target, an intermediate malicious ranking $\hat{R}^{t'}$ is constructed by manipulating the importance of these identified edges (lines 14-15). Finally, the attacker applies the internal reverse to $\hat{R}^{t'}$ to generate the final malicious ranking $\hat{R}^t$ (line 17), which amplifies the boundary gap of the poisoned global ranking.

\begin{algorithm}[tb]
    \caption{Edge Control Attack (ECA)}
    \label{alg:ECA}
    \begin{algorithmic}[1]
    \renewcommand{\algorithmicrequire}{\textbf{Input:}}
    \renewcommand{\algorithmicensure}{\textbf{Output:}}
    
    \REQUIRE Number of rounds $T$, number of malicious clients $m$, malicious clients dataset $D_m$, model weights $\theta^w$, model scores $\theta^s$, learning rate $\eta$, target accuracy $\tau$, initialization error $\varepsilon$.
        
    \ENSURE Malicious rankings $\hat{R}$.
    
    \STATE \text{ // Initialize:} $\texttt{Start} \gets \text{False},\varepsilon$
    
    \FOR{$t \in [0,T]$}
        \STATE $R_m^t \gets \text{LOCAL\_TRAINING}(D_m,\theta^w,\theta^s,\eta)$
        \IF {$ACC(R_g^{t}) \geq \tau \text{ and not } \texttt{Start}$}
            \STATE $R_\tau \gets R_g^{t}$; $\varepsilon \gets |ACC(R_g^{t}) - \tau|$; $\texttt{Start} \gets \text{True}$
        \ENDIF

        \IF{$|ACC(\hat{R}_g^t) - \tau| \leq \varepsilon \text{ and } \texttt{Start}$ }
            \STATE $R_\tau \gets \hat{R}_g^{t}$; $\varepsilon \gets |ACC(\hat{R}_g^{t}) - \tau|$
        \ENDIF
        
        \IF{$\texttt{Start}$ }
            \STATE // Manipulate AE\&DE
            \STATE $\overline{R}^t \gets \text{ESTIMATION}$
            
            \STATE $\mathrm{\mathbb{I}_{AE} , \mathbb{I}_{DE}}  \gets \text{IDENTIFY\_EDGES}(R_\tau,\overline{R}^t)$
            
            \STATE $\mathrm{{\mathbb{I}}'_{AE}},\mathrm{\mathbb{I}}'_{DE} \gets \text{MANIPULATE\_AE\&DE}(\overline{R},\mathrm{\mathbb{I}_{AE},\mathrm{\mathbb{I}_{DE}}})$
           
            \STATE $\hat{R}^{t’} \gets \text{GET\_MALICIOUS}(\overline{R},\mathrm{{\mathbb{I}}'_{AE}},\mathrm{{\mathbb{I}}'_{DE}})$
    
            \STATE // Freeze the Mask
            
            \STATE $\hat{R}^t \gets \text{INTERNAL\_REVERSE}(\mathrm{{\mathbb{I}}'_{AE}},\mathrm{{\mathbb{I}}'_{DE}},\hat{R}^{t’})$
            \ENDIF
        \ENDFOR
    \end{algorithmic}
\end{algorithm}

\begin{table*}
\caption{Comparison of RRA and our attack ECA under various AGRs in a control scenario. The ``no attack'' row reports the
global model accuracy without attack $ACC(R_g)$ (\%); other rows report control error $\xi$ (\%) under various AGRs and $\tau$.}
\vspace{-0.1cm}
\renewcommand{\arraystretch}{1.2}
\centering
\setlength{\tabcolsep}{3pt}
\resizebox{\textwidth}{!}{
\begin{tabular}{c c c c c c c c c c c c c}
\toprule
\makecell[c]{Dataset\\(Model)} & \makecell[c]{Attack} & \makecell[c]{Target\\$\tau$ (\%)} & \makecell[c]{FRL} & \makecell[c]{Multi-\\Krum-FRL} & \makecell[c]{AFA-FRL} & \makecell[c]{Fang-\\ERR-FRL} & \makecell[c]{Fang-\\LFR-FRL} & \makecell[c]{Fang-\\Union-FRL} & \makecell[c]{FABA-FRL} & \makecell[c]{DnC-FRL} & \makecell[c]{FLTrust-FRL} & \makecell[c]{Average\\$\xi$}\\
\midrule

\multirow{7}{*}{\makecell[c]{CIFAR10\\(Conv8)}} 
& no attack & / & 83.85 & 84.43 & 84.47 & 84.64 & 85.16 & 84.83 & 84.53 & 84.44 & 84.62 & / \\
\cmidrule{2-13}
& \multirow{3}{*}{\makecell[c]{RRA}} 
& 70 & 0.47 ($\pm$1.24) & 0.64 ($\pm$1.17) & 1.56 ($\pm$1.16) & 9.67 ($\pm$3.70) & 7.87 ($\pm$4.87) & 8.49 ($\pm$3.61) & 2.26 ($\pm$2.71) & 0.88 ($\pm$1.27) & 3.76 ($\pm$0.04)  & 3.96\\
& & 75 & 0.93 ($\pm$0.85) & 1.31 ($\pm$1.03) & 1.50 ($\pm$1.61) & 1.56 ($\pm$0.80) & 1.68 ($\pm$1.77) & 1.13 ($\pm$1.33) & 1.59 ($\pm$1.34) & 1.05 ($\pm$1.61) & 1.09 ($\pm$0.98) & 1.32\\
& & 80 & 0.34 ($\pm$0.69) & 0.53 ($\pm$0.59) & 0.74 ($\pm$0.53) & 2.01 ($\pm$1.37) & 2.31 ($\pm$1.45) & 1.39 ($\pm$1.01) & 2.44 ($\pm$1.80) & 0.86 ($\pm$0.74) & 1.25 ($\pm$0.81) & 1.32\\
\cmidrule{2-13}
& \multirow{3}{*}{\makecell[c]{ECA}} 
& 70 & \textbf{0.11 ($\pm$0.27)} & \textbf{0.14 ($\pm$0.28)} & \textbf{0.14 ($\pm$0.25)} & \textbf{0.11 ($\pm$0.21)} & \textbf{0.06 ($\pm$0.19)} & \textbf{0.30 ($\pm$0.21)} & \textbf{0.32 ($\pm$0.25)} & \textbf{0.34 ($\pm$0.53)} & \textbf{0.16 ($\pm$0.25)} & \textbf{0.19}\\
& & 75 & \textbf{0.02 ($\pm$0.07)} & \textbf{0.04 ($\pm$0.16)} & \textbf{0.01 ($\pm$0.04)} & \textbf{0.11 ($\pm$0.09)} & \textbf{0.05 ($\pm$0.16)} & \textbf{0.02 ($\pm$0.09)} & \textbf{0.05 ($\pm$0.16)} & \textbf{0.26 ($\pm$0.39)} & \textbf{0.06 ($\pm$0.14)} & \textbf{0.07}\\
& & 80 & \textbf{0.00 ($\pm$0.02)} & \textbf{0.02 ($\pm$0.09)} & \textbf{0.01 ($\pm$0.03)} & \textbf{0.01 ($\pm$0.04)} & \textbf{0.01 ($\pm$0.03)} & \textbf{0.01 ($\pm$0.05)} & \textbf{0.03 ($\pm$0.08)} & \textbf{0.08 ($\pm$0.16)} & \textbf{0.02 ($\pm$0.11)} & \textbf{0.02}\\
\midrule

\multirow{7}{*}{\makecell[c]{EMNIST\\(LeNet)}}
& no attack & / & 86.69 & 86.64 & 86.67 & 86.81 & 86.91 & 86.69 & 86.69 & 86.91 & 86.84 & / \\
\cmidrule{2-13}
& \multirow{3}{*}{\makecell[c]{RRA}}
& 70 & 9.13 ($\pm$1.75) & 7.73 ($\pm$2.20) & 8.15 ($\pm$1.51) & 11.50 ($\pm$1.83) & 11.43 ($\pm$2.00) & 11.23 ($\pm$1.95) & 4.56 ($\pm$3.25) & 10.98 ($\pm$2.06) & 8.55 ($\pm$1.87) & 9.25\\
& & 75 & 4.23 ($\pm$1.56) & 3.40 ($\pm$1.61) & 4.54 ($\pm$1.44) & 6.43 ($\pm$1.55) & 6.76 ($\pm$1.60) & 6.64 ($\pm$1.42) & 0.83 ($\pm$3.17) & 6.61 ($\pm$1.48) & 3.59 ($\pm$1.62) & 4.78\\
& & 80 & 0.62 ($\pm$0.93) & 0.63 ($\pm$1.11) & 0.68 ($\pm$0.98) & 2.00 ($\pm$1.00) & 1.96 ($\pm$1.16) & 1.75 ($\pm$1.04) & 0.58 ($\pm$2.16) & 1.95 ($\pm$0.97) & 0.16 ($\pm$1.13) & 1.15\\
\cmidrule{2-13}
& \multirow{3}{*}{\makecell[c]{ECA}}
& 70 & \textbf{0.47 ($\pm$0.61)} & \textbf{0.59 ($\pm$0.64)} & \textbf{0.46 ($\pm$0.52)} & \textbf{0.37 ($\pm$0.50)} & \textbf{0.44 ($\pm$0.58)} & \textbf{0.24 ($\pm$0.31)} & \textbf{0.24 ($\pm$0.46)} & \textbf{0.98 ($\pm$0.83)} & \textbf{0.23 ($\pm$0.45)} & \textbf{0.45} \\
& & 75 & \textbf{0.29 ($\pm$0.31)} & \textbf{0.26 ($\pm$0.30)} & \textbf{0.43 ($\pm$0.44)} & \textbf{0.21 ($\pm$0.27)} & \textbf{0.34 ($\pm$0.37)} & \textbf{0.13 ($\pm$0.30)} & \textbf{0.21 ($\pm$0.33)} & \textbf{0.72 ($\pm$0.62)} & \textbf{0.16 ($\pm$0.26)} & \textbf{0.31} \\
& & 80 & \textbf{0.08 ($\pm$0.17)} & \textbf{0.07 ($\pm$0.17)} & \textbf{0.02 ($\pm$0.13)} & \textbf{0.05 ($\pm$0.17) }& \textbf{0.06 ($\pm$0.15)} & \textbf{0.05 ($\pm$0.16)} & \textbf{0.05 ($\pm$0.18)} & \textbf{0.15 ($\pm$0.21)} & \textbf{0.10 ($\pm$0.17)} & \textbf{0.07} \\
\midrule

\multirow{7}{*}{\makecell[c]{FashionMNIST\\(LeNet)}}
& no attack & / & 91.35 & 91.44 & 91.30 & 91.42 & 91.33 & 91.33 & 91.33 & 91.33 & 91.33 & / \\
\cmidrule{2-13}
& \multirow{3}{*}{\makecell[c]{RRA}}
& 75 & 9.90 ($\pm$1.33) & 8.72 ($\pm$1.63) & 9.60 ($\pm$1.34) & 11.24 ($\pm$1.56) & 11.12 ($\pm$1.55) & 11.19 ($\pm$1.57) & 6.78 ($\pm$2.52) & 11.58 ($\pm$1.52) & 9.32 ($\pm$1.36) & 8.69 \\
& & 80 & 4.91 ($\pm$1.14) & 4.12 ($\pm$1.32) & 4.85 ($\pm$1.25) & 6.35 ($\pm$1.28) & 6.28 ($\pm$1.30) & 6.57 ($\pm$1.18) & 2.19 ($\pm$2.34) & 6.56 ($\pm$1.25) & 4.28 ($\pm$1.38) & 5.12 \\
& & 85 & 0.49 ($\pm$0.84) & 0.39 ($\pm$0.85) & 0.56 ($\pm$0.86) & 1.61 ($\pm$0.73) & 1.60 ($\pm$0.73) & 1.71 ($\pm$0.74) & 0.45 ($\pm$2.16) & 2.17 ($\pm$0.82) & 0.25 ($\pm$0.95) & 1.03 \\
\cmidrule{2-13}
& \multirow{3}{*}{\makecell[c]{ECA}}
& 75 & \textbf{0.40 ($\pm$0.70)} & \textbf{0.68 ($\pm$1.16)} & \textbf{0.76 ($\pm$1.14)} & \textbf{0.84 ($\pm$1.04)} & \textbf{0.47 ($\pm$0.77)} & \textbf{0.41 ($\pm$0.68)} & \textbf{0.47 ($\pm$0.86)} & \textbf{0.87 ($\pm$1.98)} & \textbf{0.86 ($\pm$1.39)} & \textbf{0.64} \\
& & 80 & \textbf{0.33 ($\pm$0.53)} & \textbf{0.10 ($\pm$0.42)} & \textbf{0.09 ($\pm$0.25)} & \textbf{0.25 ($\pm$0.46)} & \textbf{0.05 ($\pm$0.18)} & \textbf{0.13 ($\pm$0.30)} & \textbf{0.33 ($\pm$0.30)} & \textbf{0.57 ($\pm$0.86)} & \textbf{0.21 ($\pm$0.32)} & \textbf{0.23 }\\
& & 85 & \textbf{0.01 ($\pm$0.09)} & \textbf{0.02 ($\pm$0.11)} & \textbf{0.00 ($\pm$0.09)} & \textbf{0.02 ($\pm$0.09)} & \textbf{0.08 ($\pm$0.13)} & \textbf{0.05 ($\pm$0.15)} & \textbf{0.07 ($\pm$0.14)} & \textbf{0.09 ($\pm$0.25)} & \textbf{0.08 ($\pm$0.18)} & \textbf{0.05}\\
\bottomrule
\end{tabular}
}
\label{table_comparison}
\end{table*}

\section{Theoretical Analysis}
\label{throretial_analysis}
\begin{figure}[t]
  \centering
  \begin{minipage}[t]{0.48\linewidth}
      \centering
      \includegraphics[width=\linewidth]{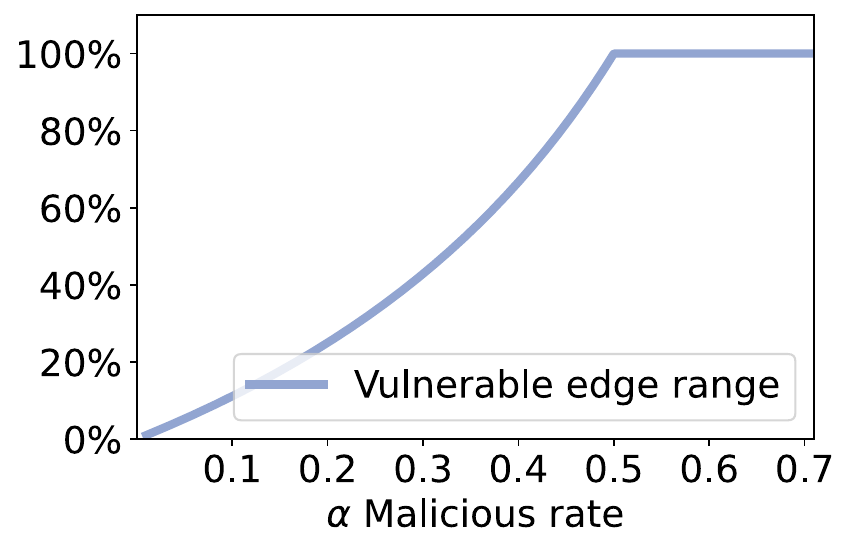}
      \caption{Vulnerable edge percentage under varying malicious rates.}
      \Description{Plot.}
      \label{fig:range}
  \end{minipage}
  \hfill 
  \begin{minipage}[t]{0.48\linewidth}
      \centering
      \includegraphics[width=\linewidth]{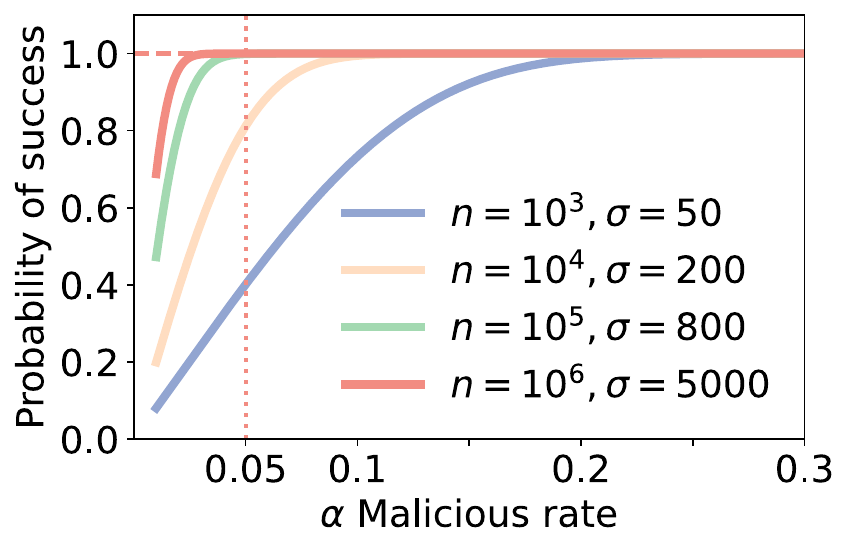}
      \caption{Success probability of edge manipulation (AE \& DE).}
      \Description{Plot.}
      \label{fig:theorem}
  \end{minipage}
\end{figure}

In this section, we theoretically analyze the range of vulnerable edges and the probability that AE and DE fall into this range, which reflects the success rate of our attack. 
Intuitively, we call an edge $e$ vulnerable if the aggregated importance score $S$ between the edge $e$ and the subnetwork boundary edge $e_k$ is less than the maximum damage that the attacker can possibly inflict in a round (either by increasing the importance score of DE or decreasing the importance score of AE). Based on the range of vulnerable edges, we can calculate the success probability of manipulating AE\&DE. Formally, we have the following theorems.

\begin{lemma}{\textbf{Vulnerable edge range.}}
\label{eq:lemma1}
Given the subnetwork percentage $k$, malicious rate $\alpha$, and the number of edges in each layer $n$, 
let $\pi^{-1}(e)$ denote the position (i.e., rank index) of edge $e$ in $R$, 
Then, the position of an edge $e$ that can be manipulated by the ECA attack is bounded by 
\begin{align}
\frac{kn-\alpha (n-1)}{(1-\alpha )}\le \pi^{-1}(e) <\frac{kn}{(1-\alpha )}.
\end{align}
\end{lemma}
Next, we apply Lemma \ref{eq:lemma1} to quantify the likelihood of AE and DE falling within the vulnerable edge range, which theoretically verifies the effectiveness of our attack.

\begin{theorem}{\textbf{Probability of Manipulating AE\&DE.}}
Let $e$ be a random edge sampled from $\mathbb{I}_{AE} \cup \mathbb{I}_{DE}$, we assume the distribution of $\pi^{-1}(e)$ approximates a discrete Gaussian distribution centered at \( \mu \) with standard deviation \( \sigma \), i.e., 
$\pi^{-1}(e) \sim D_{\mathbb{Z}, \mu, \sigma}$
. The probability of $e$ falling into the vulnerable range is 
\begin{align}
    P & \approx \Phi(\frac{2kn +(1-\alpha )(1-2\mu)}{2(1-\alpha)\sigma}) \\ \nonumber
    & \qquad - \Phi(\frac{2kn-2\alpha (n-1)-(1-\alpha )(1+2\mu)}{2(1-\alpha)\sigma}), 
\end{align} 
where $\Phi(·)$ is the cumulative distribution function of the standard normal distribution $N(0,1)$.
    \label{eq:theorem2}
\end{theorem}

We will discuss the implications of these two theorems and leave their proofs to the Appendix \ref{proof}.

\noindent\textbf{Implications.} 
Lemma~\ref{eq:lemma1} shows that the attacker's ability to manipulate edges is limited to a specific range, with only those edges within this range being controllable in a given round.  
As shown in Fig. \ref{fig:range}, increasing the malicious rate \( \alpha \) expands the vulnerable edge range, when $\alpha$ reaches 50\%, the attacker can control 100\% of the edges.
Thm.~\ref{eq:theorem2} quantifies the probability that the AE or DE falls within the vulnerable range, which reflects the success rate of our attack. 
We report the probability for the worst-case edge, i.e., the one furthest from the selection boundary, to demonstrate the lower bound of success.
As shown in Fig. \ref{fig:theorem}, where $n$ is the number of edges, $k = 50\%$, $\mu = kn$ and $\sigma$ is the standard deviation, we observe that the success probability increases as the malicious rate $\alpha$ rises. 
Even when $\alpha = 5\%$, the success probability exceeds 40\%, highlighting the threat posed by a small malicious rate. 
Additionally, when $\alpha$ is around 10\%-20\%  (depending on $n$ and $\sigma$ ), the success probability nearly reaches 100\%.

\begin{figure*}[t!]
        \begin{minipage}{0.245\linewidth}
            \centerline{\includegraphics[width=0.85\textwidth]{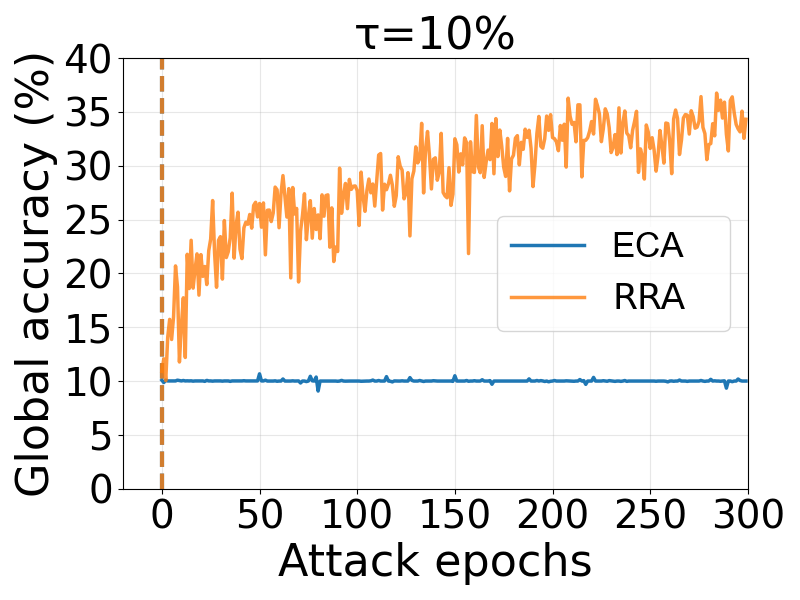}}
            \centerline{\includegraphics[width=0.85\textwidth]{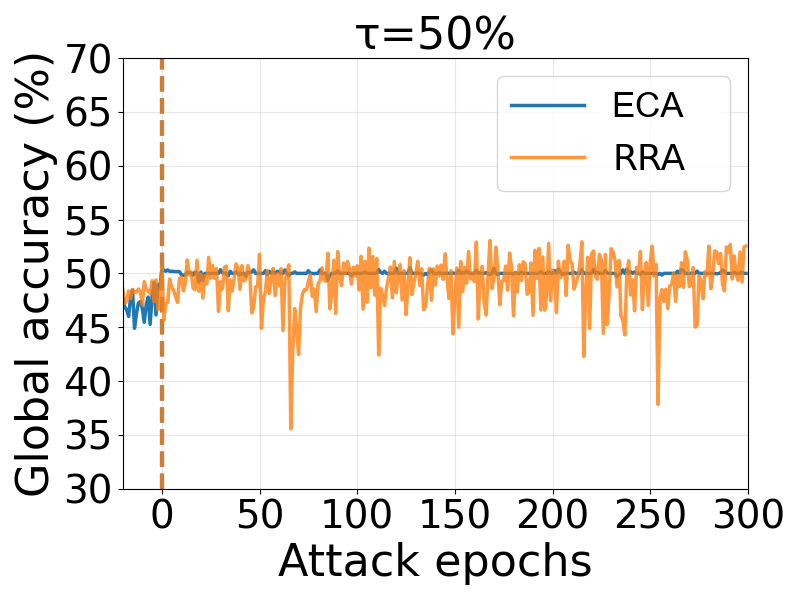}}
	\end{minipage}
        \begin{minipage}{0.245\linewidth}
		\centerline{\includegraphics[width=0.85\textwidth]{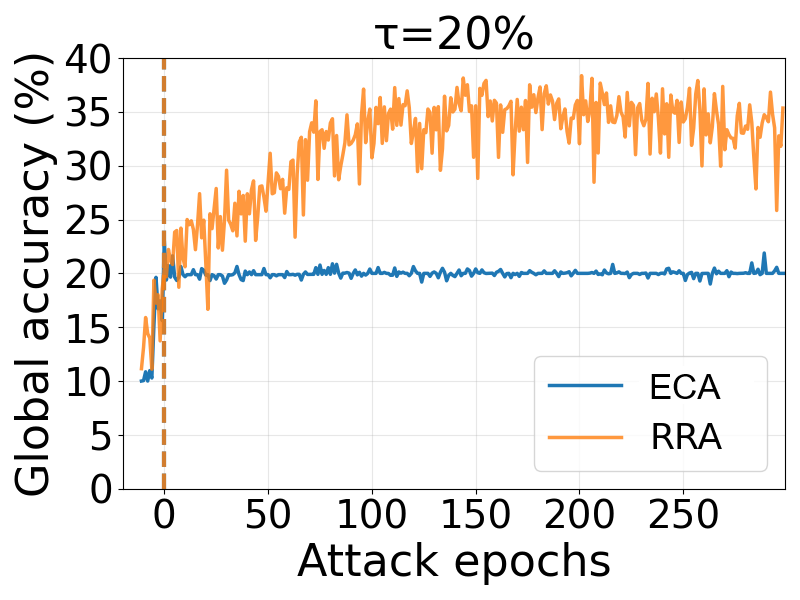}}
            \centerline{\includegraphics[width=0.85\textwidth]{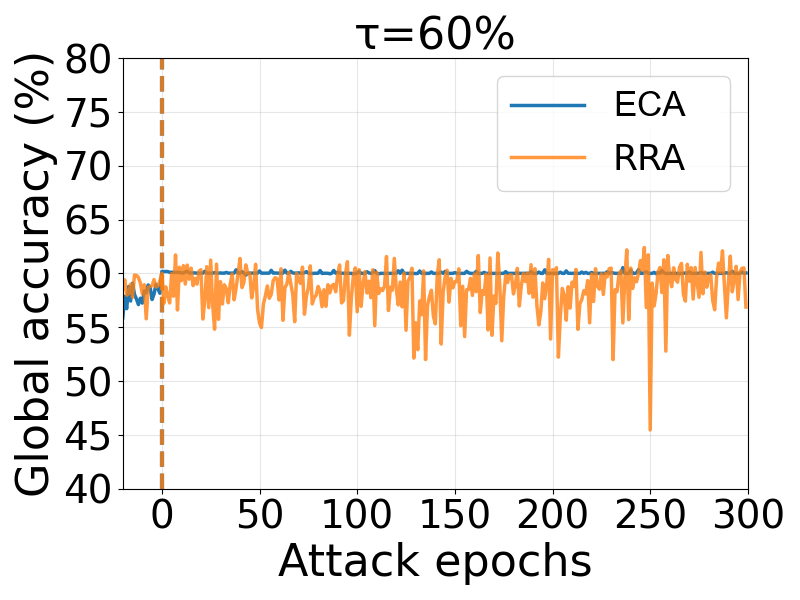}}
	\end{minipage}
        \begin{minipage}{0.245\linewidth}
            \centerline{\includegraphics[width=0.85\textwidth]{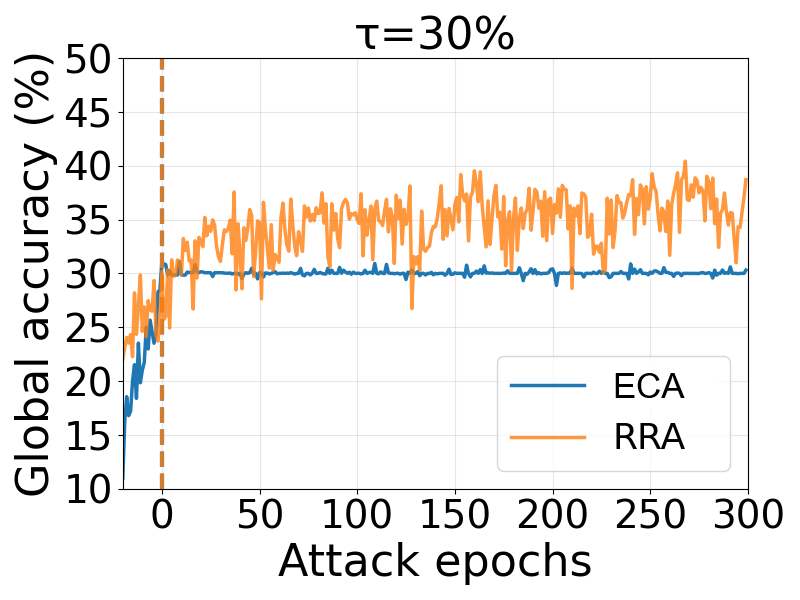}}
            \centerline{\includegraphics[width=0.85\textwidth]{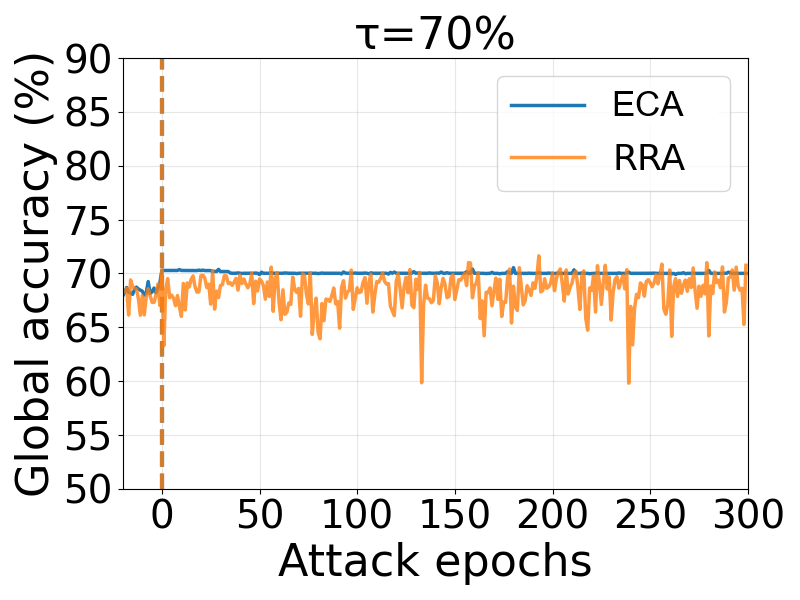}}
	\end{minipage}
        \begin{minipage}{0.245\linewidth}
            \centerline{\includegraphics[width=0.85\textwidth]{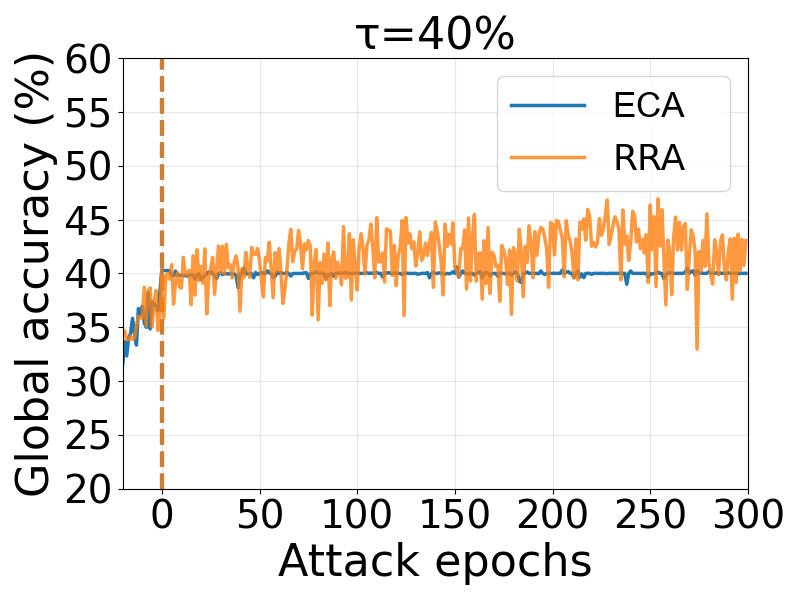}}
            \centerline{\includegraphics[width=0.85\textwidth]{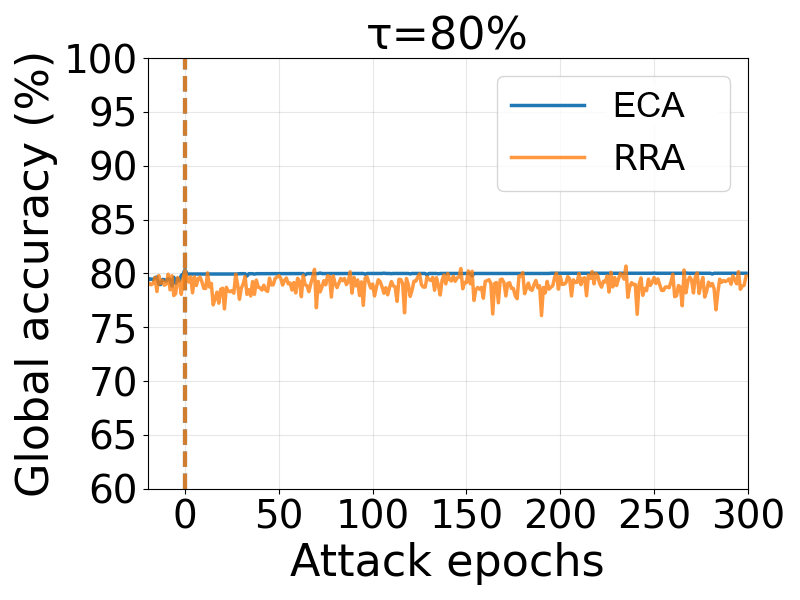}}
	\end{minipage}

    \caption{Global model accuracy (\%) under ECA and RRA attacks with different $\tau$ on the CIFAR10 dataset. The orange dash indicates the epoch when we start the attack. Note that we have only visualized the accuracy changes after the attack began.}
    \Description{Plot.}
    \label{fig:lower-tau}
    \vspace{-0.1cm}
\end{figure*}

\section{Experimental Results and Analyses}
\label{experomental_result}
\subsection{Experimental Setup}
\label{ex_setting}
\noindent\textbf{Datasets and Models.} We evaluate our attacks under seven benchmark datasets and three models. The main experiment involves CIFAR10 \citep{CIFAR10} with Conv8, EMNIST \citep{EMNIST} with LeNet, and FashionMNIST \citep{Fashion-MNIST} also with LeNet. We also describe additional datasets such as CIFAR100 \citep{CIFAR10}, Location30 \citep{location30_1,location30_2}, Purchase100 \citep{Purchase100}, and Texas100 \citep{L&P&Tdata} in the Appendix \ref{experiment-details}.

\noindent\textbf{FL Settings.} We consider the cross-device FL setting, where 1000 clients collaborate to train a global model, with the server randomly selects 25 clients each training round. We run our attack for 500 rounds after the global model reaches the target accuracy $\tau$. For our main experiment, we adopt a more realistic scenario by distributing the data in a non-IID manner using the Dirichlet distribution with parameter $\beta$ = 1. Unless otherwise stated, we assume a $20\%$ malicious rate.
 
\noindent\textbf{Poisoning Attacks for Comparison.} Our goal is to precisely control the accuracy of the global model to a given target value $\tau$ under the FRL framework.
However existing fine-grained model control attacks \citep{zhang2023denial} do not apply to FRL, therefore we propose a possible attack called Random Ranking Attack (RRA) for comparison. 
Specifically, in RRA, malicious clients upload randomized rankings once the model accuracy exceeds the target threshold $\tau$, causing slight performance degradation. Conversely, when the model accuracy drops below $\tau$, the attacker ceases the attack, allowing the model to naturally recover. This strategy causes the global model's performance to oscillate around the target accuracy $\tau$, achieving a certain degree of control without fully disrupting the training process.

\noindent\textbf{Evaluated Defenses.} We consider nine SOTA AGRs integrated with FRL. Not only including distance-based (Multi-Krum \citep{blanchard2017machine}, AFA \citep{AFA}, FABA \citep{FABA},DnC \citep{shejwalkar2021manipulating}), weighted-aggregation (FLTrust \citep{cao2020fltrust}) and validation-based AGRs (Fang \citep{fang2020local}). 
Notably, we do not consider norm-bounded \citep{sun2019can} and dimension-wise AGRs \citep{yin2018byzantine} because these approaches are incompatible with the FRL. More details about AGRs are deferred to Appendix \ref{experiment-details}.

\noindent\textbf{Evaluation Metrics.} Let $\tau$ denote the target global accuracy for the attacker, and $ACC(\hat{R}_g)$ denote the global model accuracy after the attack. We use control error, i.e., $\xi = |ACC(\hat{R}_g) - \tau|$ to evaluate the performance of the control attack. A smaller $\xi$ indicates a more successful attack.

\subsection{Evaluation Results}
Our extensive experiments confirm the potent effectiveness and precision of ECA. As shown in Table. \ref{table_comparison}, ECA demonstrates a high level of control across all combinations of datasets, target accuracy, and defense methods, with an average control error of only  0.224\% and is 17 $\times$ smaller than RRA.
Specifically, under the FRL framework, the average control error is 0.19\% (18.1 $\times$ lower compared to RRA). In the most robust case DnC-FRL, the average control error is 0.45\% (10.5 $\times$ lower compared to RRA). More experimental results are provided in Appendix~\ref{additional_experiment}, including: computational efficiency analysis; studies on additional datasets (CIFAR100, Location30, Purchase100, Texas100); evaluations against SOTA defenses (FLCert, FoundationFL); and performance under lower target accuracies ($\tau$) and minimal malicious rates (1\%).

We also visualize the control performance of ECA and RRA. Fig.~\ref{fig:lower-tau} shows that RRA exhibits significant fluctuations, while ECA demonstrates a smoother and more precise control. Notably, RRA struggles to control the global model accuracy at low $\tau$ values, such as $\tau=10\%$ and $\tau=20\%$, where the global accuracy exhibits large fluctuations and deviates significantly from the target value. In contrast, our ECA consistently maintains the global model accuracy close to the target with minimal variance. This stark difference highlights the superiority of ECA, particularly in complex settings where fine-grained control is critical, making it more reliable than RRA.

\subsection{Ablation Studies}
\begin{figure}
    \centering
    \includegraphics[width=1\linewidth]{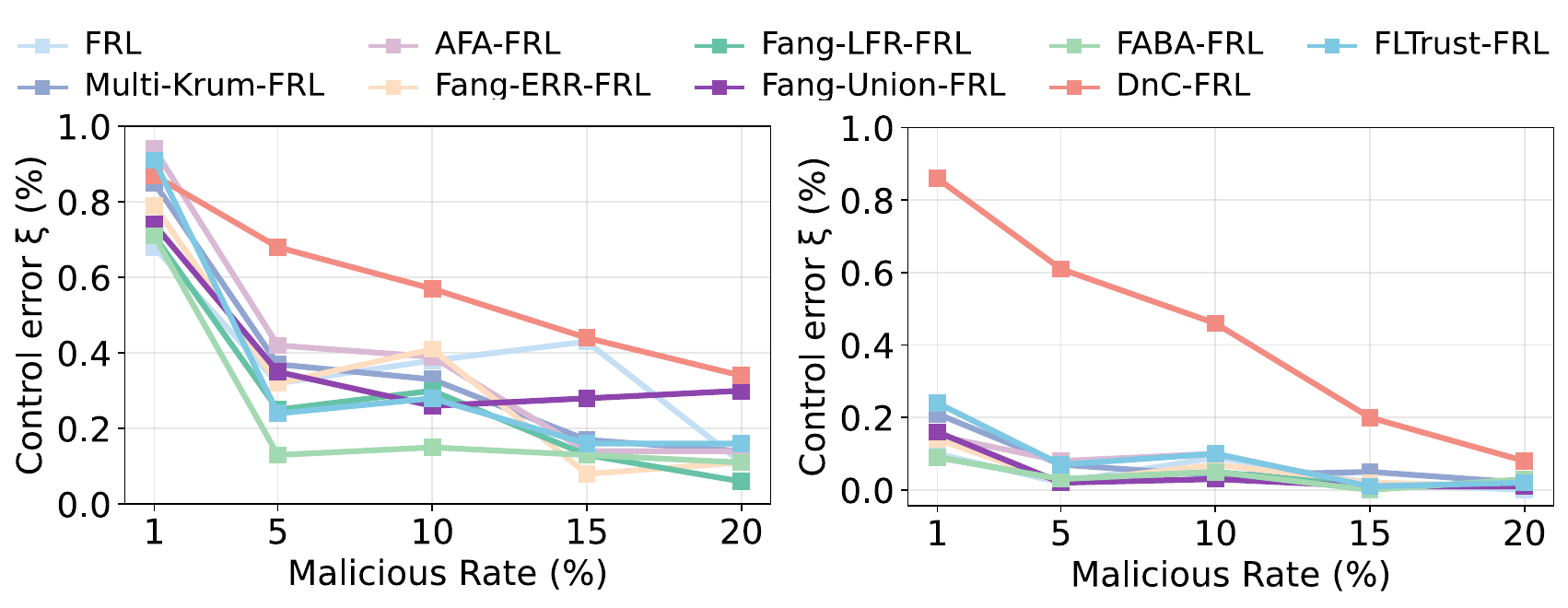}
    \caption{Control errors $\xi$ (\%)  for different defenses at $\tau$ = 70\% (left) and $\tau$ = 80\% (right) with different malicious rates on the CIFAR10 dataset.}
    \Description{Plot.}
    \label{fig:mal_rate}
    \vspace{-0.1cm}
\end{figure}

\noindent\textbf{Impact of the malicious rate.}
Fig.~\ref{fig:mal_rate} shows the ECA attack results when the malicious rate changes from 1\% to 20\% under the FRL framework. We notice that as the proportion of malicious clients increases, the control effect becomes better. 
Notably, even with only 1\% of malicious clients, the attack achieves the desired control effect, with an average control error of 0.823\%.

\begin{figure}
    \centering
    \includegraphics[width=\linewidth]{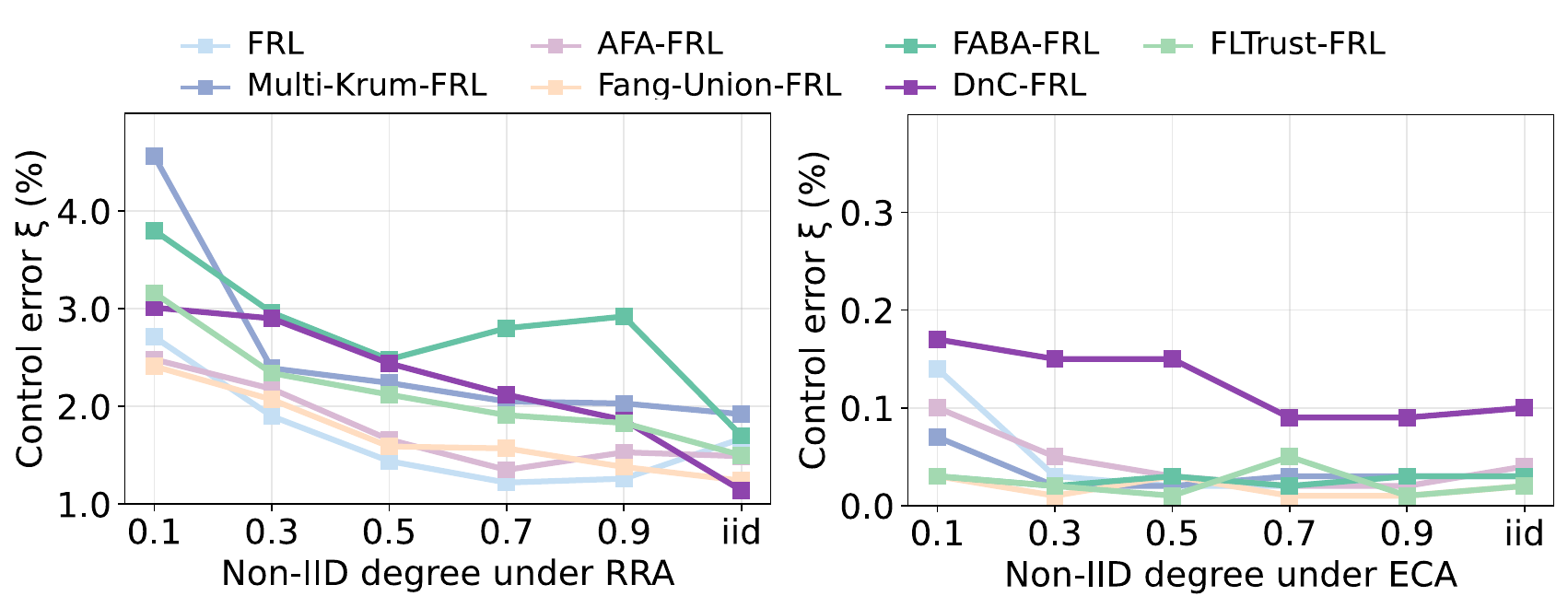}
    \caption{Control error $\xi$ (\%) for different defenses on different non-iid degree $\beta$ (x-axis) when $\tau$ = 70\% on the CIFAR10 dataset.}
    \Description{Plot.}
    \label{fig:iid}
    \vspace{-0.1cm}
\end{figure}

\noindent\textbf{Impact of the Non-IID degree.}
Fig.~\ref{fig:iid} illustrates the impact of varying non-IID degrees on the control error. The hyperparameter $\beta$ defines the degree of non-IID, where smaller $\beta$ values indicate a higher non-IID degree. 
As $\beta$ decreases, the data distribution becomes increasingly heterogeneous, making it more challenging to predict benign client updates. Despite this, our approach consistently achieves the desired attack performance.
Specifically, our ECA attack achieves a control error of less than 0.2\%, the average control error is 0.045\%, significantly outperforming RRA by 47.3$\times$. 

\begin{figure}
    \centering
    \includegraphics[width=\linewidth]{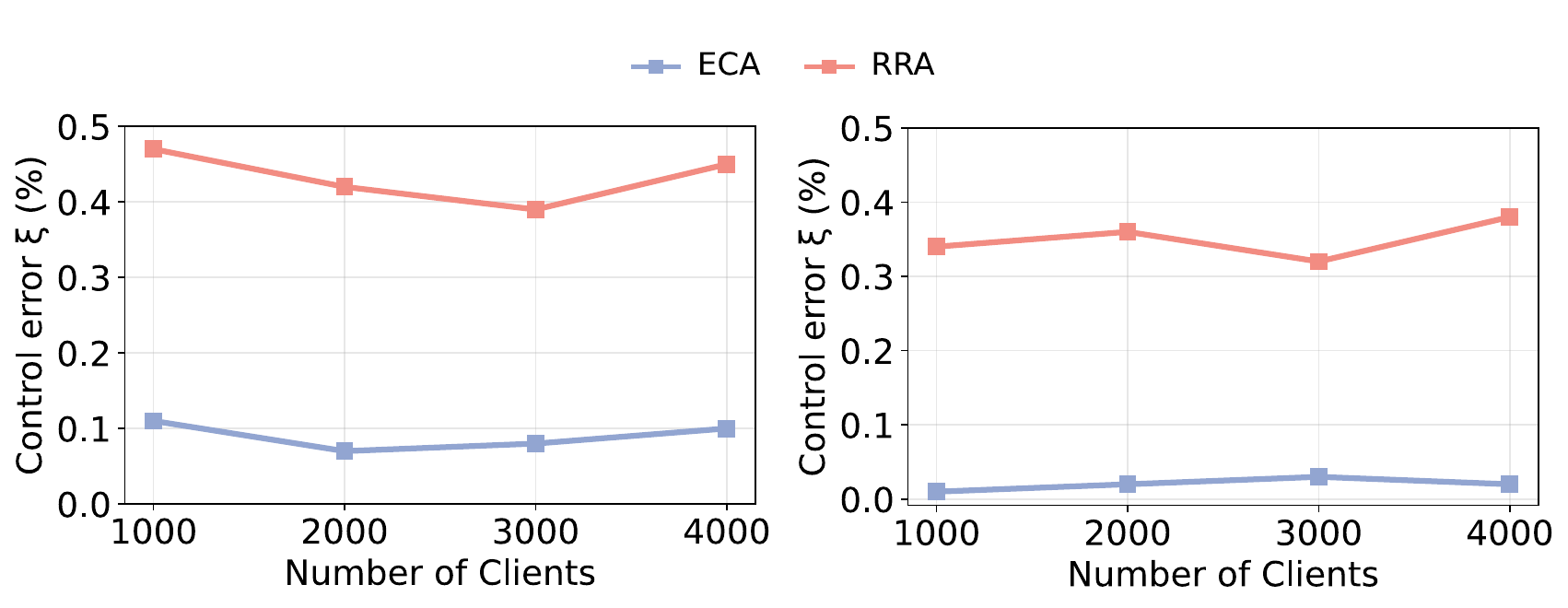}
    \caption{Control error $\xi$ (\%) for different numbers of clients under $\tau$ = 70\% (left) and $\tau$ = 80\% (right) on the CIFAR10 dataset.}
    \Description{Plot.}
    \label{fig:clients}
    \vspace{-0.1cm}
\end{figure}

\noindent\textbf{Impact of number of clients.}
Fig.~\ref{fig:clients} illustrates the impact of the number of total clients on the control error for ECA and RRA attacks under FRL. The results show that the number of clients has almost no effect on ECA.

\begin{figure}
    \centering
    \includegraphics[width=0.9\linewidth]{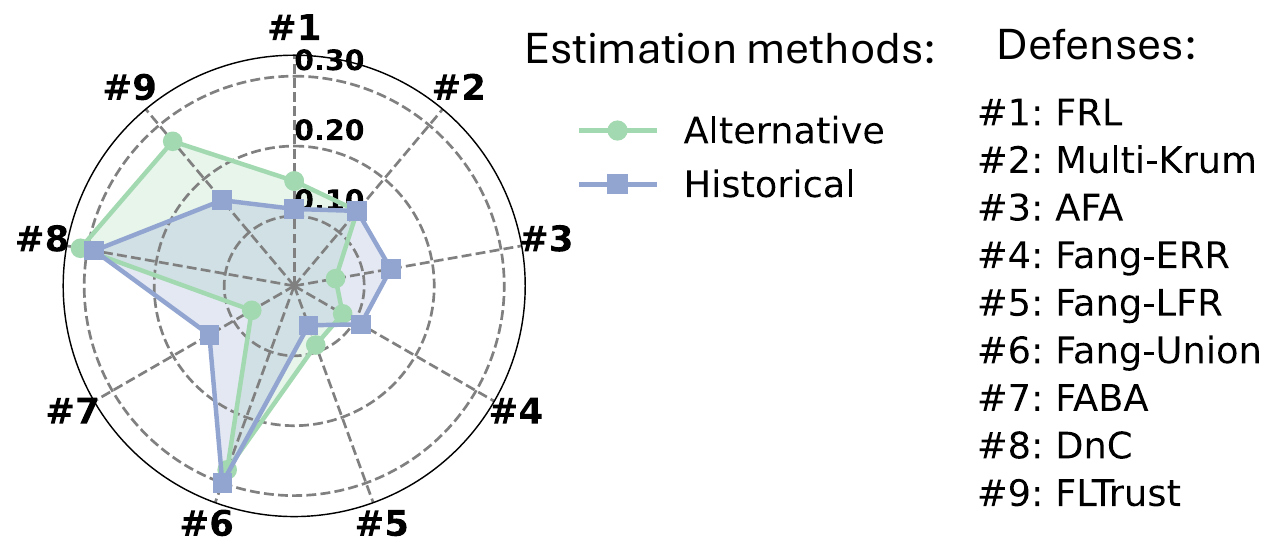}
    \caption{Control errors $\xi$ (\%) for different estimation methods under $\tau $ = 70\% on the CIFAR10 dataset.}
    \Description{Plot.}
    \label{fig:estimate}
    \vspace{-0.1cm}
\end{figure}

\noindent\textbf{Impact of the estimation method.}
We compare control errors from two estimation methods: Historical and Alternative estimation. As shown in Fig.~\ref{fig:estimate}, both methods achieve the desired performance, with control errors below 0.35\%. However, the Historical estimation method demonstrates slightly superior performance, making it the preferred choice for our main experiments.

\section{Discussion}
\paragraph{Comparison with Related Work.} We now provide a detailed comparison with the state-of-the-art MPA designed for FRL, namely VEM \cite{SPzirui}, from three key perspectives: attack objective, technical methodology, and empirical performance.
The main difference between VEM and ECA lies in their fundamental attack objectives. VEM aims for a Denial of Service (DoS) attack, which focuses on causing a continuous degradation of global model performance without any bounds. In contrast, ECA introduces a novel fine-grained control attack. This attack is both dynamic and bidirectional, designed to maintain model performance precisely at a specified target value. Its objective is to prevent the performance from exceeding or falling below the target accuracy, making it a significantly greater challenge. As a result, existing optimization-based methods are not suitable for addressing this type of attack.

The differences in objectives require fundamentally different technical approaches. VEM employs a stochastic optimization method, specifically the Gumbel-Sinkhorn method, to iteratively search for a maximally damaging update. In contrast, ECA utilizes a deterministic two-stage algorithm that directly constructs the malicious ranking. This distinction also extends to the types of vulnerability that are exploited. In VEM, edges are static and defined globally based on their proximity to the mask boundary. On the other hand, ECA's Ascending and Descending Edges (AE/DE) are task-specific and determined dynamically. Our experiments reveal a significant \textbf{96\% non-overlap} between these two sets of edges.

\begin{figure}
    \centering
    \includegraphics[width=\linewidth]{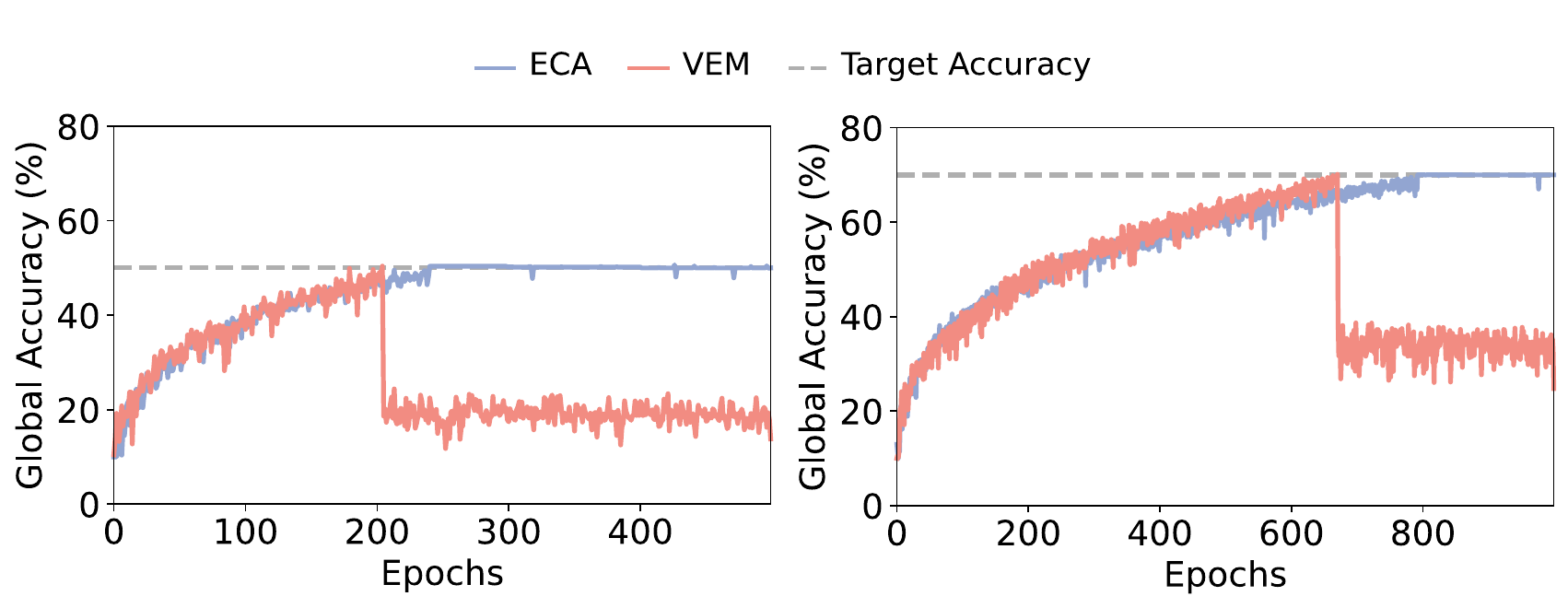}
    \caption{Convergence curves of ECA and VEM attacks on the CIFAR10 dataset when $\tau$ = 50\% (left) and $\tau$ = 70\% (right).}
    \label{fig:compare}
\end{figure}

Empirically, Table~\ref{tab:vem_vs_eca_comparison} shows VEM's control error grows with target accuracy, whereas ECA maintains consistently low error. Fig.~\ref{fig:compare} further confirms ECA's stable convergence compared to VEM's volatility. This demonstrates that fine-grained control is a distinct challenge requiring our unique algorithmic solution.

\begin{table}[h!]
\centering
\caption{Comparison of the control error $\xi$ (\%) of our method and VEM under the FashionMNIST dataset, LeNet architecture, and different target accuracies $\tau$ 
 (\%).}
\label{tab:vem_vs_eca_comparison}
\begin{tabular}{lccccc}
\toprule
\textbf{Attack} & \textbf{$\tau=50\%$} & \textbf{$\tau=55\%$} & \textbf{$\tau=60\%$} & \textbf{$\tau=65\%$} & \textbf{$\tau=70\%$} \\
\midrule
VEM& 5.92& 3.51& 5.13& 9.71& 14.72\\
\textbf{ECA (Ours)}& \textbf{1.95}& \textbf{1.18}& \textbf{0.87}& \textbf{0.59}& \textbf{0.76}\\
\bottomrule
\end{tabular}
\end{table}

\paragraph{Limitations and Future Work.} Our work currently focuses on untargeted attacks; extending it to targeted scenarios is a priority for future work. We will also explore its effectiveness against more robust FL frameworks and advanced defenses. The broader impact of our work underscores the critical need to evaluate the security of promising FL frameworks like FRL, which is based on the LTH and aims for lightweight models. Ultimately, these explorations highlight the urgent need to develop more resilient FL methods, particularly for enhancing the security of discrete-space FL paradigms.

\section{Conclusion}
\label{conclusion}
In this work, we reassess security analysis for ranking-based FL, revealing its vulnerability to sophisticated threats beyond simple disruptions. 
Our analysis shows that an adversary can exploit FRL's update mechanism not just to damage a model but to precisely steer its accuracy to a specific, attacker-chosen level. This capability is particularly dangerous in competitive commercial scenarios, where an attacker can stealthily degrades a competitor's model to secure a decisive strategic advantage without raising alarms.
Based on the observations, we proposed the ECA, the first fine-grained control attack target FRL system. Our novel two-stage algorithm successfully achieves the attacker's dual objectives of precise control and stealth. We provided a theoretical foundation for ECA, formally characterizing the attack's feasibility and establishing the precise conditions and high probability of its success. Our extensive experiments then confirmed this theoretical promise, demonstrating that ECA can consistently maintain control with an exceptionally low average error of 0.224\% across diverse settings. This confirms that the strategic threat we identified is not merely theoretical but highly practical and effective.

\section{Ethical Considerations}
This research reveals FRL vulnerabilities strictly to improve defense. Experiments rely exclusively on public datasets and open-source models, involving no personal data. We advocate responsible disclosure and will release code with appropriate safeguards.

\section*{Acknowledgments}
This work is supported in part by the Australian Research Council under grant LP240100315.

\bibliography{www2026_arxiv}
\bibliographystyle{ACM-Reference-Format}

\clearpage
\appendix

\section{Proof}
\label{proof}
\subsection{Proof of Lemma 1.}
In this section, we detail the proof of Lemma \ref{eq:lemma1}. We will prove the upper bound of the vulnerable edge position $\pi^{-1}(e_{mal})$, and the method of proving the lower bound will be similar.

Since the aggregated importance scores of the benign clients $S_{ben}$ is an increasing sequence, we assume that the $S_{ben}$ contribution from benign clients is $S_{ben}(e) = \sum_{u=1}^{U-m} I_u(e) \approx (U-m) \pi^{-1}(e)$.

We consider the edge ($e_{ae} \in \mathrm{\mathbb{I}_{AE}}$) to be manipulated that has the largest benign position in $\mathrm{\mathbb{I}_{AE}}$, and in order to place this edge below the boundaries of the final poisoned global model $\hat{R}_g$, the ECA method gives this edge the lowest score in the malicious rankings, that is, $I(e_{ae})=0$. So we have the following equation:
\begin{align}
    &\sum_{u= 1}^{U-m}I_u(e_{ae}) + \sum _{v= 1}^{m} I_v(e_{ae}) < \sum _{z= 1}^{U} I_z(e_{k}) \nonumber \\
    & \approx (U-m)\times \pi^{-1}(e_{ae})+m\times 0 < U\times kn \nonumber \\
    & = (1-\alpha)U\pi^{-1}(e_{ae})<knU \nonumber \\
    & = \pi^{-1}(e_{ae})< \frac{kn}{(1-\alpha)},
    \label{eq:upper-bound}
\end{align}
where $e_k$ denotes the smallest edge at the boundary, i.e., the edge at the $kn^{th}$ position.

Similarly, we consider the edge $e_{de}$ to be manipulated to have the least benign position in $\mathrm{\mathbb{I}_{DE}}$, and in order to place this edge above the boundaries of the final poisoned global model $\hat{R}_g$, the ECA method gives the highest score to this edge in the malicious rankings, i.e., $I(e_{de})=n-1$. Thus, we have the following equation:
\begin{align}
    &\sum_{u= 1}^{U-m}I_u(e_{de}) + \sum _{v= 1}^{m} I_v(e_{de}) \ge \sum _{z= 1}^{U} I_z(e_{k}) \nonumber \\
    & \approx (U-m)\times \pi^{-1}(e_{de})+m\times (n-1) \ge U\times kn \nonumber \\
    & = (1-\alpha)U\pi^{-1}(e_{de}) +\alpha U(n-1) \ge knU \nonumber \\
    & = \pi^{-1}(e_{de}) \ge \frac{kn-\alpha (n-1)}{(1-\alpha )}.
    \label{eq:lower-bound}
\end{align}

Given all the $\mathrm{\mathbb{I}_{AE}}$ and $\mathrm{\mathbb{I}_{DE}}$, we obtain the range of benign positions $\pi^{-1}(e_{mal})$ that the attacker can control by considering their worst-case scenarios in combination with Eq. \ref{eq:upper-bound} and Eq. \ref{eq:lower-bound}, we have:
\begin{align}
\frac{kn-\alpha (n-1)}{(1-\alpha )}\le \pi^{-1}(e_{mal}) <\frac{kn}{(1-\alpha )}.
\end{align}

\subsection{Proof of Theorem 1.}
In this section, we present a detailed proof of Theorem \ref{eq:theorem2}. Additionally, we provide our experimental results, which confirm the effectiveness of our attack with higher true probabilities compared to the theoretical values.

\begin{figure}[t!]
    \centering
    \begin{minipage}{0.48\linewidth}
        \centering
        \includegraphics[width=\textwidth]{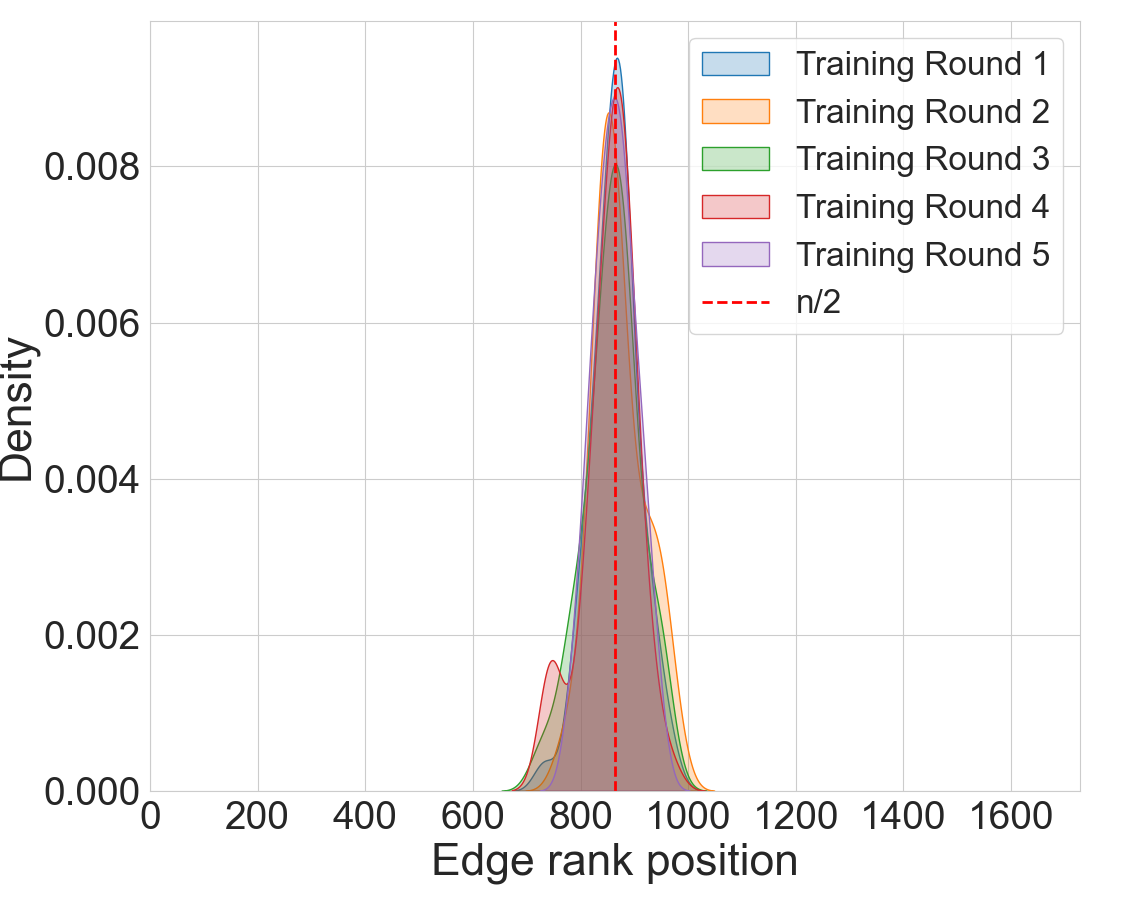}
        \caption{Kernel density distribution of $\mathrm{\mathbb{I}_{AE}}\cup \mathrm{\mathbb{I}_{DE}}$ position data.}
        \Description{Plot.}
        \label{fig:data_distribution}
    \end{minipage}
    \vspace{-0.2cm}
    \hfill
    \begin{minipage}{0.48\linewidth}
        \centering
        \includegraphics[width=\textwidth]{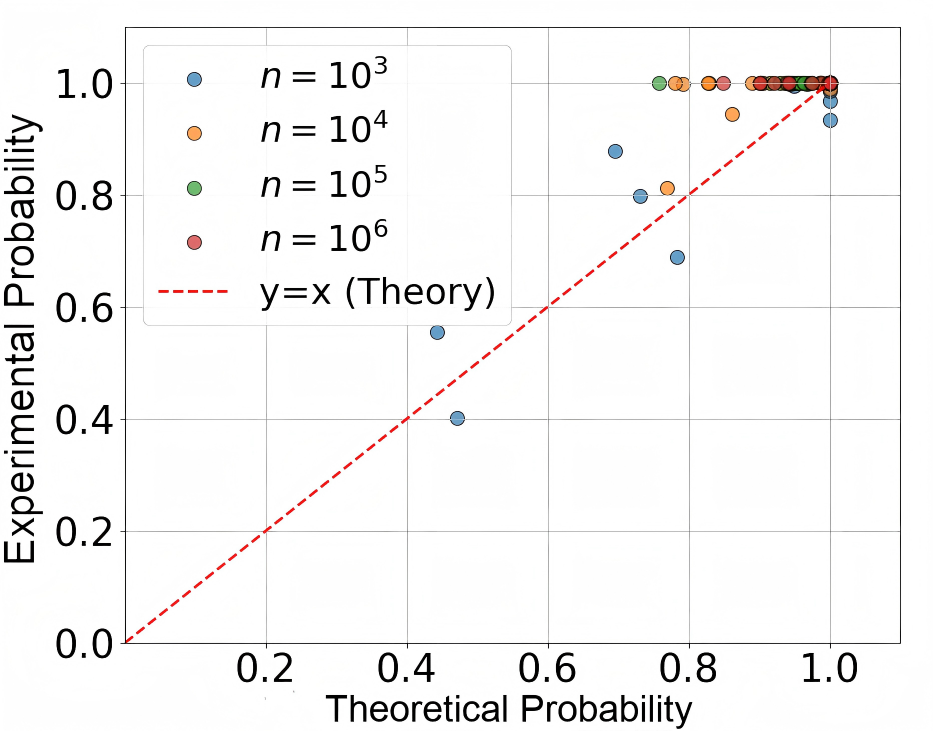}
        \caption{Comparison of theoretical and experimental probabilities.}
        \Description{Plot.}
        \label{fig:scatter}
    \end{minipage}
    \vspace{-0.2cm}
\end{figure}

When performing an ECA attack, the attacker selects a set of target edges ($\mathrm{\mathbb{I}_{AE}}\cup \mathrm{\mathbb{I}_{DE}}$). Given these edges, we treat the positions of the edges as a random variable. We assume that the positional distribution approximates a discrete Gaussian distribution centered at \( \mu \) with standard deviation \( \sigma \), i.e., $\pi^{-1}(e) \sim D_{\mathbb{Z}, \mu, \sigma}$.
Fig.~\ref{fig:data_distribution} shows that we conducted multiple rounds of experimentation, resulting in similar distributions for the position data. The red dashed line indicates the $kn$ positions with \( k = 50\% \). We found that the kernel density distribution of \(\mathrm{\mathbb{I}_{AE}} \cup \mathrm{\mathbb{I}_{DE}}\) overlapped significantly across the rounds, with the highest density region concentrated around \(\frac{n}{2}\).
This is highly consistent with our intuition about the scheme, since the edges at the boundary are prone to change, so we consider the expectation of this distribution to be $\mu = kn$, that is, $\pi^{-1}(e) \sim D_{\mathbb{Z}, kn, \sigma}$.

According to Lemma \ref{eq:lemma1}, we get the range of vulnerable edges that the attacker can control, so the probability that the attacker can successfully control the DE and AE edges can be defined as the probability that any of the target edges fall into the vulnerable range. We use the successive Gaussian probabilities within the interval as the discrete probabilities, i.e.,
\begin{equation}
P(X = k) = \Phi\left(\frac{k+0.5-\mu}{\sigma}\right)-\Phi\left(\frac{k-0.5-\mu}{\sigma}\right),
\end{equation}
for interval $[k-0.5,k+0.5)$.
To simplify the notation, let us define the lower and upper bounds of the vulnerable range as $L$ and $U$, respectively:
\[
L = \frac{kn-\alpha (n-1)}{1-\alpha}, \quad U = \frac{kn}{1-\alpha}.
\]
So we have:
\begin{align}
P
& = P( L \le \pi ^{-1}(e) < U ), \nonumber \\
& = {\textstyle \sum_{z=\lfloor L \rfloor}^{\lceil U \rceil}} P(\pi ^{-1}(e) = z), \nonumber\\
& \approx \underline{\Phi((L+0.5-\mu)/\sigma )} - \Phi((L-0.5-\mu)/\sigma ) \nonumber\\
& \quad + \underline{\Phi((L+1+0.5-\mu)/\sigma )} - \underline{\Phi((L+1-0.5-\mu)/\sigma )} \nonumber\\
& \quad + \cdots \nonumber\\
& \quad + \Phi((U+0.5 -\mu)/\sigma ) - \underline{\Phi((U-0.5 -\mu)/\sigma )}, \nonumber\\
& \approx  \Phi((U+0.5 -\mu)/\sigma ) - \Phi((L-0.5-\mu)/\sigma ), \nonumber\\
& \approx \Phi\left(\frac{2kn +(1-\alpha )(1-2\mu)}{2(1-\alpha)\sigma}\right) \nonumber \\
& \qquad - \Phi\left(\frac{2kn-2\alpha (n-1)-(1-\alpha )(1+2\mu)}{2(1-\alpha)\sigma}\right).
\end{align}

As displayed in Fig. \ref{fig:scatter}, we conducted experiments at various edge scales ranging from \(10^3\) to \(10^6\) to assess the likelihood of a real attacker successfully manipulating AE\&DE. In each of these experiments, we observed the success rates over 20 trials. Our findings revealed that the majority of experimental success probabilities exceeded those predicted by the theoretical model. Additionally, as the size of \(n\) increased, our experimental probabilities often reached 100\%, which further validates the effectiveness of our attacks.

\section{Details of Experiment Settings}
\label{experiment-details}
In this section, we provide the details of the experimental setup that are missing from Section \ref{ex_setting}.

\subsection{Datasets}
\label{datasets}
For each dataset experiment, we train the model using a default sparsity rate of 0.5. In local training, we follow the configuration outlined in \cite{mozaffari2023every}: each client trains for 5 epochs using the SGD optimizer with the following parameters: learning rate 0.4, momentum 0.9, weight decay 1e-4, and batch size 32.

\noindent\textbf{CIFAR10} \cite{CIFAR10} is a 10-class classification task consisting of 60000 RGB images (50,000 for training and 10,000 for testing), each sized by 32×32. We experiment with Conv8 architecture. 

\noindent\textbf{EMNIST} \cite{EMNIST} consists of 814255 gray-scale images of numeric, upper, and lower case handwritten letters, with a total of 62 categories. We experiment with the LeNet architecture. 

\noindent\textbf{Fashion-MNIST} \cite{Fashion-MNIST} is a 10-class fashion image for classification tasks, using a training set of 60000 images and a testing set of 10000 images with a resolution of 28×28 pixels. We experiment with the LeNet architecture. 

\noindent\textbf{CIFAR100} \cite{CIFAR10} is a standard 100-class image classification benchmark, consisting of 60,000 32×32 RGB images (50,000 for training and 10,000 for testing), with 500 training and 100 test samples per class. We experiment with Conv8 architecture. 

\noindent\textbf{Location30} \cite{location30_1,location30_2} is a 30-class geosocial classification dataset derived from the Foursquare check-in records, containing 5,010 samples with 446 binary features. Each feature indicates whether a user has visited a specific region or location type. We obtain a simplified and preprocessed Location dataset provided by Shokri et al. \cite{L&P&Tdata}. We experiment with FC architecture. 

\noindent\textbf{Purchase100} \cite{Purchase100} is a 100-class classification task comprising 197,324 data samples with 600 binary features, derived from Kaggle's Acquire Valued Shoppers Challenge and preprocessed by Shokri et al. \cite{L&P&Tdata}. Each feature indicates the purchase status of a specific product, and samples are categorized into distinct consumer behavior patterns. We experiment with FC architecture. 

\noindent\textbf{Texas100} \cite{L&P&Tdata} is a 100-class medical procedure prediction dataset derived from public patient records released by the Texas Department of State Health Services. The preprocessed dataset \cite{L&P&Tdata} contains 67,330 samples with 6,170 binary characteristics, encoding patient demographics (e.g., age, sex), external causes of injury (e.g., suicide, drug misuse) and diagnoses (e.g., schizophrenia). The classification task predicts the most frequent procedures (e.g., surgery) from patient profiles. We experiment with FC architecture. 

\subsection{Evaluated Defenses and Settings}
\label{evaluated_defenses}
\noindent\textbf{Krum} \cite{blanchard2017machine} selects a single local update as the global model by identifying the update closest (in Euclidean distance) to $U-m-2$ neighboring updates, where $m$ is the upper bound on the number of malicious clients, and $U$ is the total number of selected clients in the current training round.

\noindent\textbf{Multi-Krum} \cite{blanchard2017machine} extends Krum by selecting $c$ local updates such that $U-c>2*m+2$. These $c$ updates are then averaged to form the global model.

\noindent\textbf{Adaptive federated average (AFA)} \cite{AFA} computes cosine similarity between each local update and a benign reference gradient. Gradients with similarities falling outside a predefined range—determined by the mean, median, and standard deviation of all similarities—are discarded.

\noindent\textbf{Fang} \cite{fang2020local} assumes that the server has access to a clean validation dataset and uses it to evaluate the uploaded model updates for detecting malicious clients. This method consists of three variants: Fang-ERR, Fang-LFR, and Fang-Union. Fang-ERR identifies malicious updates by assessing the classification error rate on the validation dataset, where updates leading to significantly higher error rates are flagged as malicious. Fang-LFR, on the other hand, evaluates the loss values of the updated models on the validation dataset, marking updates with substantially higher losses as potential outliers. Fang-Union combines the two approaches by simultaneously considering both error rates and loss values; an update is deemed malicious only if it exhibits abnormal behavior in both metrics, thereby enhancing the robustness and accuracy of the detection process. This methodology effectively leverages the server's clean validation dataset to distinguish between benign and malicious updates in federated learning frameworks.

\noindent\textbf{FABA} \cite{FABA} iteratively removes the local update that is furthest from the average of all updates until the number of excluded updates equals the estimated number of attackers.

\noindent\textbf{Divide-and-conquer (DnC)} \cite{shejwalkar2021manipulating} uses singular value decomposition (SVD) to detect and filter outliers. It first creates a gradient subsample by randomly selecting indices from the input gradient dimensions, then centers the subsample by subtracting its mean. The centered gradients are projected onto the top singular eigenvector, capturing the dominant variation direction. Gradients with the highest projection scores are removed as outliers. 

\noindent\textbf{FLTrust} \cite{cao2020fltrust} incorporates a trusted validation dataset on the server to compute reference gradients. Each client is assigned a trust score based on the similarity between its local update and the reference gradient. These trust scores are used as weights when aggregating to form the global model.

\begin{table*}
\centering
\caption{Comparison of RRA and our attack ECA under various AGRs in a control scenario. The ``no attack'' row reports the
global model accuracy without attack $ACC(R_g)$ (\%); other rows report control error $\xi$ (\%) under various AGRs and $\tau$.}
\vspace{-0.2cm}
\footnotesize
\resizebox{\textwidth}{!}{
\begin{tabular}{c c c c c c c c c c c c}
\toprule
\makecell[c]{Dataset\\(Model)} & \makecell[c]{Attack} & \makecell[c]{Target\\$\tau$ (\%)} & \makecell[c]{FRL} & \makecell[c]{Multi-\\Krum-FRL} & \makecell[c]{AFA-FRL} & \makecell[c]{Fang-\\ERR-FRL} & \makecell[c]{Fang-\\LFR-FRL} & \makecell[c]{Fang-\\Union-FRL} & \makecell[c]{FABA-FRL} & \makecell[c]{DnC-FRL} & \makecell[c]{FLTrust-FRL} \\
\midrule

\multirow{7}{*}{\makecell[c]{CIFAR10\\(Conv8)}} 
& no attack & / & 83.85 & 84.43 & 84.47 & 84.64 & 85.16 & 84.83 & 84.53 & 84.44 & 84.62 \\
\cmidrule{2-12}
& \multirow{3}{*}{\makecell[c]{RRA}} 
& 40 & 2.41 ($\pm$1.58) & 1.65 ($\pm$1.52) & 2.01 ($\pm$1.41) & 3.00 ($\pm$1.96) & 1.90 ($\pm$2.11) & 1.99 ($\pm$1.48) & 2.07 ($\pm$2.10) & 3.98 ($\pm$2.66) & 1.83 ($\pm$1.41)  \\
& & 50 & 1.72 ($\pm$1.70) & 1.99 ($\pm$1.79) & 1.62 ($\pm$1.88) & 1.73 ($\pm$1.26) & 1.84 ($\pm$1.56) & 1.70 ($\pm$1.25) & 2.10 ($\pm$2.20) & 2.39 ($\pm$1.62) & 1.75 ($\pm$1.73) \\
& & 60 & 1.68 ($\pm$1.32) & 2.23 ($\pm$2.38) & 1.63 ($\pm$1.73) & 1.39 ($\pm$1.24) & 1.65 ($\pm$1.52) & 1.77 ($\pm$1.99) & 2.25 ($\pm$2.18) & 1.57 ($\pm$1.27) & 1.77 ($\pm$1.68) \\
\cmidrule{2-12}
& \multirow{3}{*}{\makecell[c]{ECA}} 
& 40 & \textbf{0.18 ($\pm$0.27)} & \textbf{0.14 ($\pm$0.40)} & \textbf{0.11 ($\pm$0.22)} & \textbf{0.08 ($\pm$0.18)} & \textbf{0.05 ($\pm$0.10)} & \textbf{0.08 ($\pm$0.19)} & \textbf{0.17 ($\pm$0.28)} & \textbf{0.34 ($\pm$0.54)} &\textbf{ 0.09 ($\pm$0.23)} \\
& & 50 & \textbf{0.09 ($\pm$0.17)} & \textbf{0.05 ($\pm$0.13)} &\textbf{ 0.05 ($\pm$0.10)} & \textbf{0.05 ($\pm$0.10)} & \textbf{0.04 ($\pm$0.08)} & \textbf{0.09 ($\pm$0.20)} & \textbf{0.06 ($\pm$0.12)} & \textbf{0.38 ($\pm$0.58)} & \textbf{0.08 ($\pm$0.14)} \\
& & 60 & \textbf{0.10 ($\pm$0.27) }& \textbf{0.04 ($\pm$0.08)} & \textbf{0.03 ($\pm$0.09)} & \textbf{0.06 ($\pm$0.13)} & \textbf{0.04 ($\pm$0.14)} & \textbf{0.04 ($\pm$0.09)} &\textbf{ 0.03 ($\pm$0.08)} & \textbf{0.18 ($\pm$0.41)} & \textbf{0.06 ($\pm$0.12)} \\
\midrule
\end{tabular}
}
\label{table2_comparison}
\vspace{-0.3cm}
\end{table*}
\section{More Experiment Result}
\label{additional_experiment}

\noindent\textbf{Computational efficiency analysis.}
Our proposed attack is computationally efficient as its core manipulation is based on reordering operations rather than costly optimization. The process involves identifying differing edges (AE/DE), adjusting malicious rankings, and applying an internal reverse strategy. The time complexity is primarily bounded by sorting, at \textbf{$\mathcal{O}(n\log n)$}, which is necessary for generating or aggregating rankings. This contrasts sharply with optimization-based attacks that require extensive resources for iterative gradient-based solutions, often involving numerous model propagations. Our method circumvents these complex procedures by oper ating directly in the discrete ranking space, allowing for the direct construction of malicious updates.

The attack construction phase imposes a minimal computational burden on the adversary. While malicious clients still perform standard local training (Algorithm \ref{alg:ECA}, line 3), the subsequent manipulation is lightweight. This is especially true when using the "historical estimation" method, which leverages the previous round's global ranking and requires negligible additional computation. This efficiency allows the attack to be executed effectively even in resource-constrained environments, enhancing its practical threat and stealth.

\begin{table}[ht]
\centering
    \caption{Control error $\xi$ (\%) of our attack on other AGRs with different malicious rates under FRL framework.}
    \vspace{-0.2cm}
    \renewcommand{\arraystretch}{1.3}
    \resizebox{0.5\textwidth}{!}{
    \begin{tabular}{c c c  c c c}
    \toprule
    AGRs & Malicious Rate & $\tau = 70\%$ & $\tau = 75\%$ & $\tau = 80\%$ & Average $\xi$\\
    \midrule
    \multirow{2}{*}{FLCert}& 10\% & 0.43 ($\pm$0.35) & 0.15 ($\pm$0.11) & 0.11 ($\pm$0.13) &0.36 \\
     & 20\% & 0.15 ($\pm$0.27) & 0.05 ($\pm$0.10) & 0.04 ($\pm$0.08) & 0.11\\
    \midrule
    \multirow{2}{*}{FoundationFL} & 10\% & 0.55 ($\pm$0.49) & 0.31 ($\pm$0.37) & 0.23 ($\pm$0.15) & 0.40\\
    & 20\% & 0.35 ($\pm$0.29) & 0.11 ($\pm$0.25) & 0.06 ($\pm$0.19) & 0.20\\
    \bottomrule
    \end{tabular}}
    \vspace{-0.3cm}
    \label{defense}
\end{table}

\begin{table}[ht]
\caption{Average control error $\xi$ (\%) of our attack on other datasets with different malicious rates under FRL framework.}
\vspace{-0.2cm}
\centering
\renewcommand{\arraystretch}{1.1}
\resizebox{0.5\textwidth}{!}{
\begin{tabular}{c ccccc}
\hline
Malicious& CIFAR100 & Location30 & Purchase100 & Texas100 \\
  rate   & (Conv8)  & (FC)       & (FC)        & (FC)     \\ \hline
10\%     & 0.33 ($\pm$0.35) & 0.21 ($\pm$0.22) & 0.48 ($\pm$0.41) & 0.27 ($\pm$0.31)    \\
20\%     & 0.03 ($\pm$0.05) & 0.07 ($\pm$0.12) & 0.15 ($\pm$0.20) & 0.06 ($\pm$0.11)  \\ \hline
\end{tabular}
}
\vspace{-0.3cm}
\label{more_dataset}
\end{table}

\begin{figure}[H]
    \centering
    \includegraphics[width=0.9\linewidth]{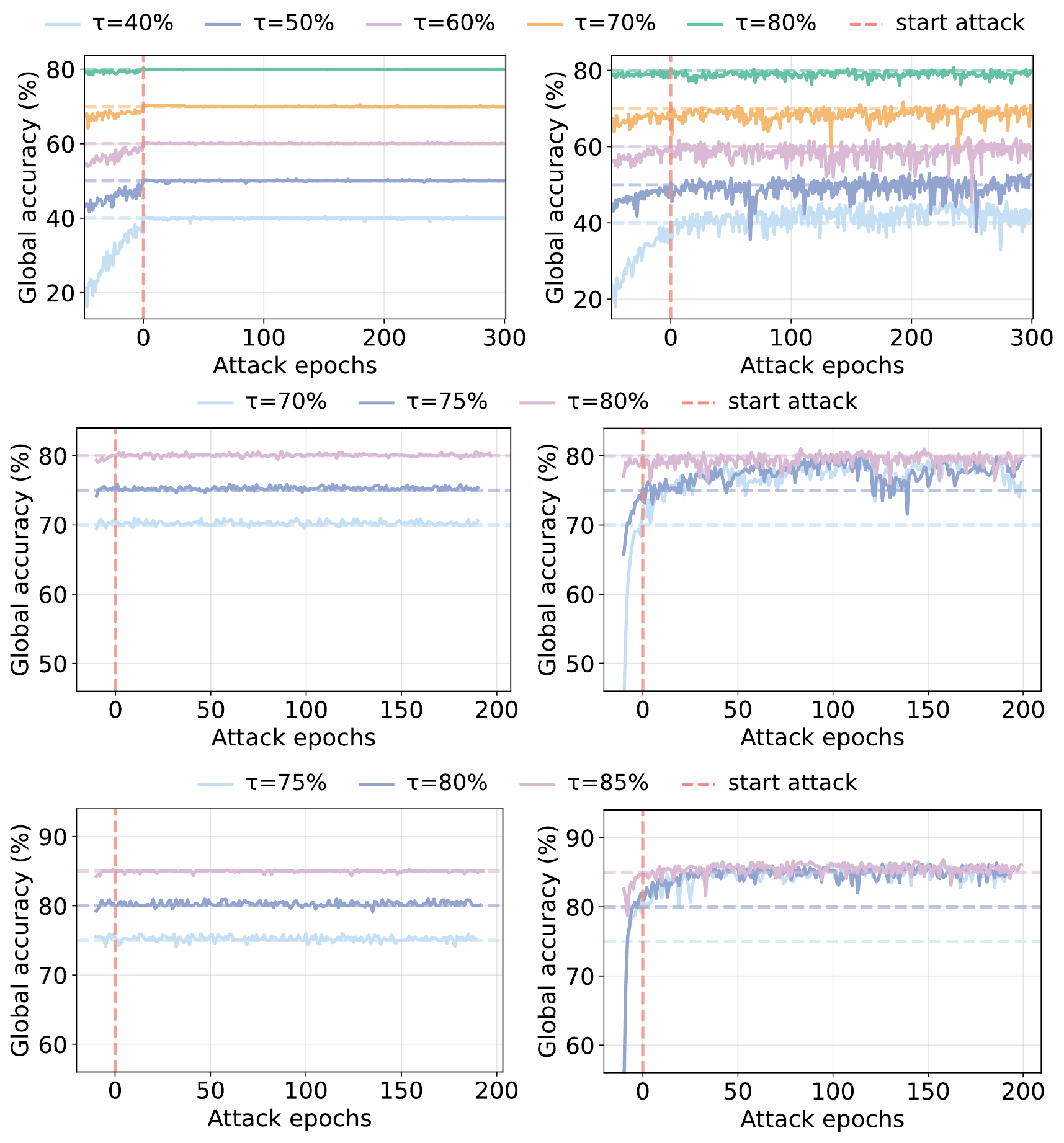}
    \caption{Global model accuracy (\%) of ECA (left) and RRA (right) on CIFAR10 (first row), EMNIST (second row), and Fashion-MNIST (third row) datasets with different $\tau$.}
    \Description{Plot.}
    \label{fig:c-ECA-tau}
    \vspace{-0.3cm}
\end{figure}

\noindent\textbf{Impact of lower $\tau$ values.}
We further compared ECA with RRA using lower target $\tau$ values on the CIFAR-10, EMNIST, and FashionMNIST datasets. As detailed in Table \ref{table2_comparison} for CIFAR-10, ECA consistently achieves superior precision with control errors below 0.4\%, significantly outperforming RRA, which exhibits much higher errors (1.57\% to 3.98\%). This quantitative advantage is visually confirmed across all three datasets in Fig.~\ref{fig:c-ECA-tau}. These figures collectively demonstrate that under ECA, the global model accuracy rapidly converges to and stabilizes at the target $\tau$ shortly after the attack begins. In stark contrast, RRA's performance shows substantial fluctuations and an inability to maintain stable control, a weakness that is particularly pronounced at lower $\tau$ values. These results underscore ECA's robustness and precision in achieving fine-grained model control, a capability where RRA is demonstrably deficient.

\noindent\textbf{Experiments on additional defense mechanisms.}
We evaluated our attack against SOTA defenses: FoundationFL \citep{FoundationFL}, an augmented defense, and FLCert \citep{cao2022flcert}, a leading authentication-based defense. As shown in Table \ref{defense}, our attack maintains precise control with low average error $\xi$ (\%) against both. For instance, with a 20\% malicious rate, the average error was merely 0.11\% against FLCert and 0.20\% against FoundationFL. These results demonstrate that our attack effectively bypasses these advanced defenses, achieving its objective with consistently low error rates regardless of the defense mechanism or malicious participation level.

\noindent\textbf{Experiments on additional datasets.} Table \ref{more_dataset} presents the average error control of our attack on larger datasets, as well as three tabular datasets with 10\% and 20\% malicious rates under the FRL framework. From the table, we can see that our attack achieves desirable attack control in all cases.

\begin{table}[h]
\centering
\caption{Control error $\xi$ (\%) of our ECA under various AGRs.}
\label{tab:low_malicious_rate_results}
\vspace{-0.2cm}
\renewcommand{\arraystretch}{1.3}
\resizebox{0.5\textwidth}{!}{
\begin{tabular}{c c c c c c c c}
\toprule
\textbf{Target $\tau$} & \textbf{Malicious Rate} & \textbf{FRL} & \textbf{FLTrust-FRL} & \textbf{FABA-FRL} & \textbf{DnC-FRL} & \textbf{Fang-Union-FRL} \\
\midrule
40(\%) & 1(\%) & 1.27 ($\pm$1.90) & 1.40 ($\pm$1.42) & 1.64 ($\pm$1.92) & 1.90 ($\pm$2.27) & 1.32 ($\pm$1.22) \\
 70(\%) & 1(\%) & 0.68 ($\pm$0.48) & 0.91 ($\pm$0.83) & 0.71 ($\pm$0.55) & 0.87 ($\pm$0.60) & 0.74 ($\pm$0.49) \\
 40(\%) & 2(\%) & 0.97 ($\pm$1.52) & 1.21 ($\pm$1.45) & 1.35 ($\pm$1.88) & 1.53 ($\pm$1.39) & 1.21 ($\pm$1.75) \\
 70(\%) & 2(\%) & 0.61 ($\pm$0.48) & 0.47 ($\pm$0.36) & 0.54 ($\pm$0.35) & 0.75 ($\pm$0.43) & 0.41 ($\pm$0.50) \\
\bottomrule
\end{tabular}
}
\vspace{-0.3cm}
\end{table}

\noindent\textbf{Impact of lower malicious rates.}
We conduct additional experiments to verify the effectiveness of our method under lower malicious user rates. As shown in Table \ref{tab:low_malicious_rate_results}, our ECA attack demonstrates potent effectiveness against a wide array of AGRs even with only 1\% malicious clients. For a control target of $\tau = 40$\%, the control error ($\xi$) remains consistently low across all defenses, such as  
$1.40\pm1.42$ against FLTrust-FRL and $1.32 \pm1.22$ against Fang-Union-FRL. Furthermore, when we increase the control target to $\tau = 70$\%, the attack's precision improves and the control error drops even further; for instance, the error against Fang-Union-FRL decreases from $1.32\pm1.22$ to $0.74\pm0.49$. This trend is amplified when the malicious client rate is increased to 2\%, where the control error for $\tau = 70$\% against Fang-Union-FRL is reduced to just $0.41\pm0.50$.

\end{document}